\documentclass[runningheads]{llncs}
\usepackage{cite}
\usepackage{abbr}
\usepackage{graphicx}
\usepackage{amsmath,amssymb} % define this before the line numbering.
\usepackage{color}
\usepackage{multirow}
\usepackage{subfigure}
\usepackage{hyperref}
\usepackage[width=122mm,left=12mm,paperwidth=146mm,height=193mm,top=12mm,paperheight=217mm]{geometry}

\renewcommand{\paragraph}{\noindent\textbf}
\begin{document}
% \linenumbers
\pagestyle{headings}
\mainmatter
\def\ECCV18SubNumber{1587}  % Insert your submission number here

\title{Adversarial Binary Coding\\ for Efficient Person Re-identification} % Replace with your title

%\titlerunning{ECCV-18 submission ID \ECCV18SubNumber}
\titlerunning{Adversarial Binary Coding for Efficient Person Re-identification}

%\authorrunning{ECCV-18 submission ID \ECCV18SubNumber}
\authorrunning{Zheng Liu, Jie Qin, Annan Li, Yunhong Wang, and Luc Van Gool}

%\author{Anonymous ECCV submission}
%\institute{Paper ID \ECCV18SubNumber}

\author{Zheng Liu\inst{1} \and Jie Qin\inst{2} \and Annan Li\inst{1} \and Yunhong Wang\inst{1} \and Luc Van Gool\inst{2}}
\institute{Beijing Advanced Innovation Center for Big Data and Brain Computing,\\ Beihang University, Beijing, China\\ \and Computer Vision Laboratory, ETH Z{\"u}rich, Switzerland %\email{\{zhengliu,liannan,yhwang\}@buaa.edu.cn}~~\email{\{jqin,vangool\}@vision.ee.ethz.ch}
}

\maketitle

\begin{abstract}
Person re-identification (ReID) aims at matching persons across different views/scenes. In addition to accuracy, the matching efficiency has received more and more attention because of demanding applications using large-scale data. Several binary coding based methods have been proposed for efficient ReID, which either learn projections to map high-dimensional features to compact binary codes, or directly adopt deep neural networks by simply inserting an additional fully-connected layer with tanh-like activations. However, the former approach requires time-consuming hand-crafted feature extraction and complicated (discrete) optimizations; the latter lacks the necessary discriminative information greatly due to the straightforward activation functions. In this paper, we propose a simple yet effective framework for efficient ReID inspired by the recent advances in adversarial learning. Specifically, instead of learning explicit projections or adding fully-connected mapping layers, the proposed Adversarial Binary Coding (ABC) framework guides the extraction of binary codes implicitly and effectively. The discriminability of the extracted codes is further enhanced by equipping the ABC with a deep triplet network for the ReID task. More importantly, the ABC and triplet network are simultaneously optimized in an end-to-end manner. Extensive experiments on three large-scale ReID benchmarks demonstrate the superiority of our approach over the state-of-the-art methods.
\end{abstract}

%%%%%%%%% BODY TEXT
\section{Introduction}
\label{sec:intro}

\begin{figure}[t]
\begin{center}

\subfigure[Learning Projections]{ \label{subfig:projections}
  \includegraphics[width=0.30\linewidth]{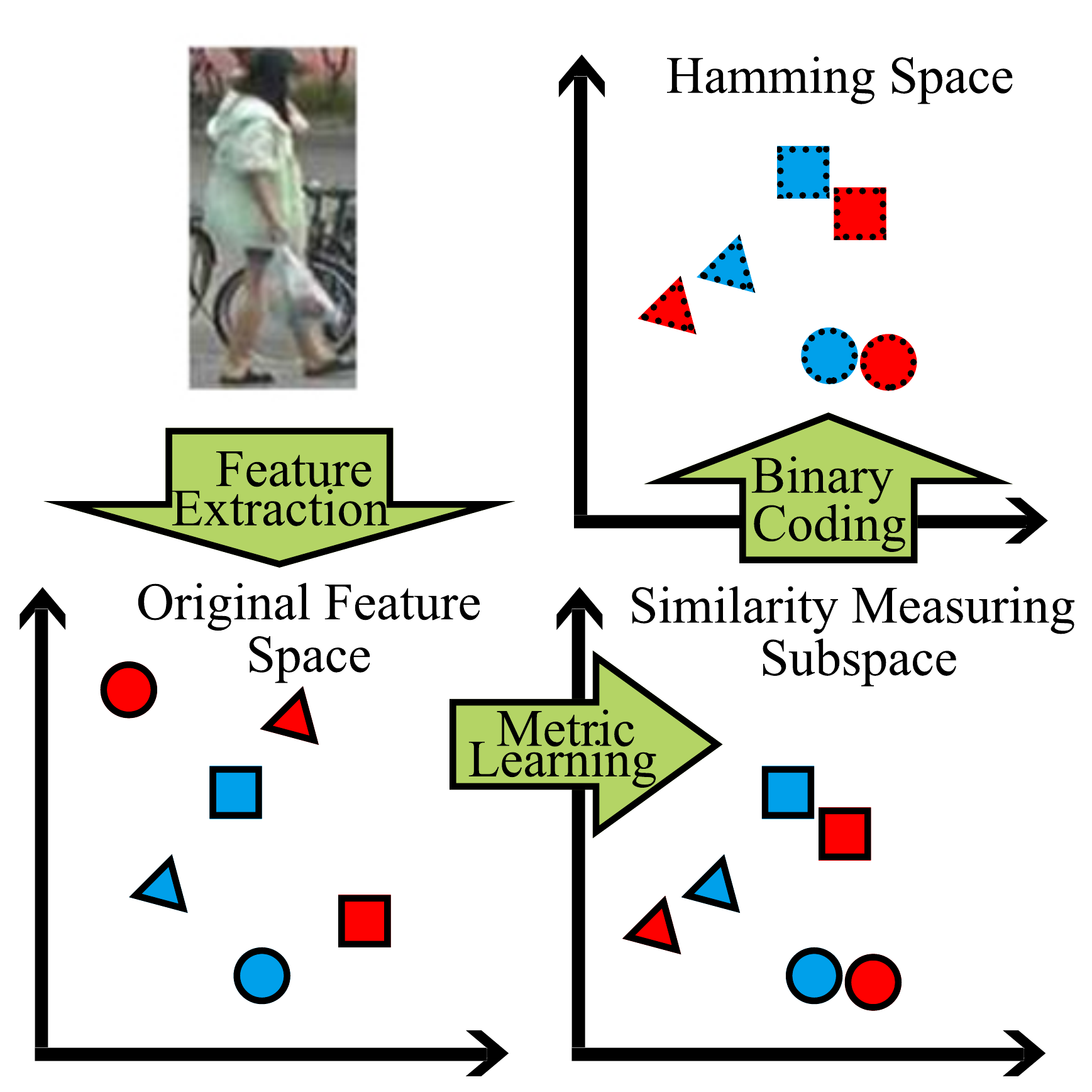}
}
\subfigure[Deep Hashing]{ \label{subfig:deephashing}
  \includegraphics[width=0.30\linewidth]{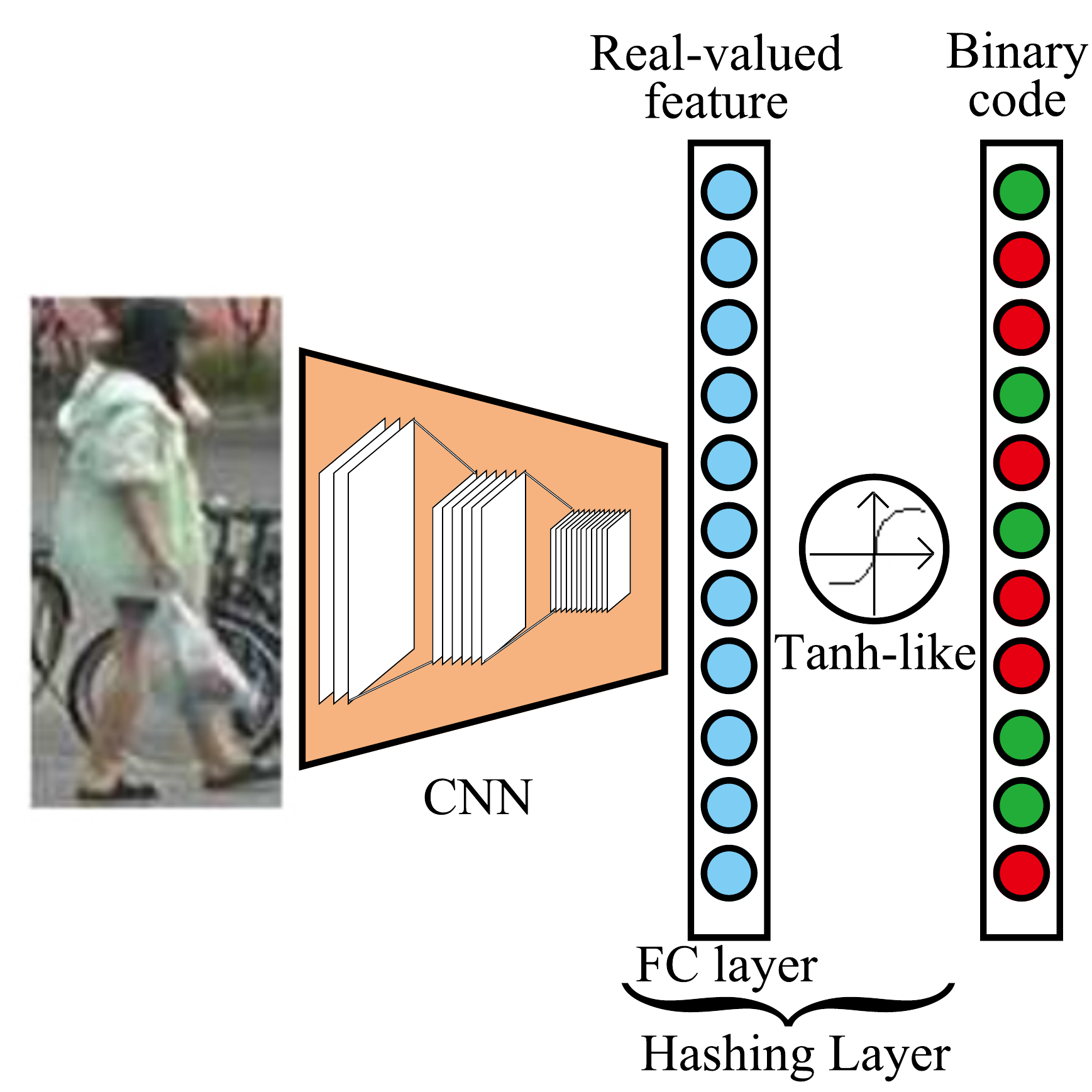}
}
\subfigure[Ours]{ \label{subfig:abc}
  \includegraphics[width=0.30\linewidth]{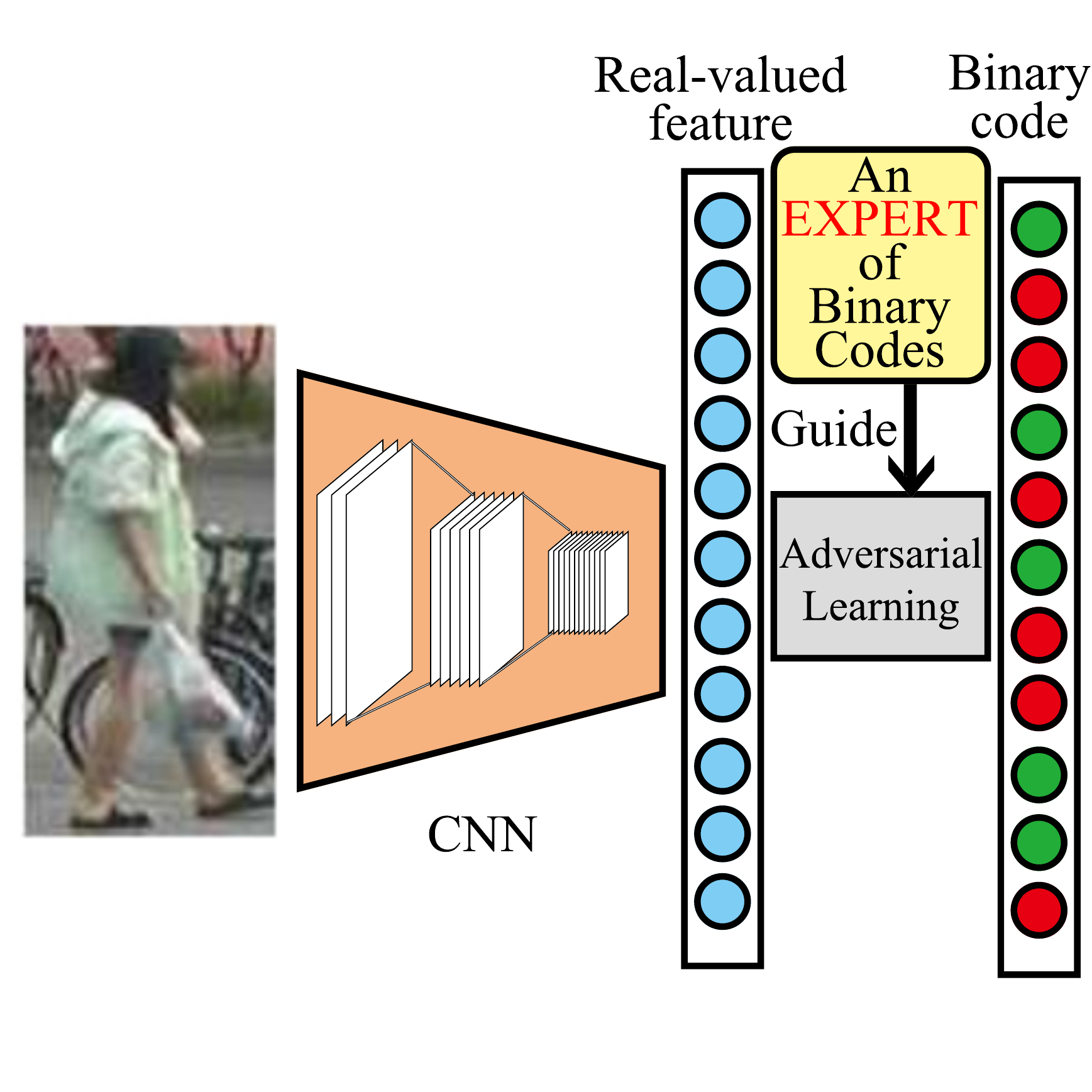}
}
\end{center}
\vspace{-3.6mm}
   \caption{Different binary coding schemes for efficient ReID. Our method avoids time-consuming projection learning and results in high-quality binary codes in an intuitive way, via adversarial training without using tanh-like activation.}
\label{fig:differenthashing}
\vspace{2mm}
\end{figure}

Given one or multiple images of a pedestrian, person re-identification (ReID) aims to retrieve the person with the same identity from a large collection of images captured in different scenes and from various viewpoints. ReID enables various potential applications, such as long-term cross-scenario tracking and criminal retrieval. The task, however, still remains challenging due to the significant variations in poses, viewpoints and illuminations across different cameras.

Numerous ReID methods have been proposed, most of which adopt high-dimensional (usually thousands or more) features \cite{farenzena2010person,liao2015person,Matsukawa_2016_CVPR,liu2015spatio,shi2015transferring,wang2016joint} in order to represent persons comprehensively with various cues (\eg colors, textures, and spatial-temporal cues). This directly bring much higher computational complexity to the subsequent similarity measurement (\eg metric learning). Besides, current large-scale ReID benchmarks contain numerous identities and cameras to simulate real-world scenarios, making existing state-of-the-art ReID approaches computationally unaffordable \cite{zheng2016person}. Therefore, despite the noteworthy improvement in matching accuracies, the computational and memory requirements have at the same time become more challenging.

%Besides, current ReID benchmarks tend to contain much more identities and cameras to simulate real-world scenarios. As a result, the time consumed by learning specific objectives during training and searching candidates during testing increases significantly. Moreover, loading tens of thousands of high-dimensional features into memory is also a heavy burden \cite{zheng2016person}. Therefore, despite the currently noteworthy success in matching accuracies, the computation and storage efficiencies have recently become much more challenging issues of ReID.

Binary coding (\ie hashing), adopted by \eg \cite{zheng2016learning,ChenJiaxin_2017_CVPR}, maps high-dimensional features into compact binary codes and efficiently measures similarities in the low-dimensional \emph{Hamming space}. It is one of the promising solutions for efficient ReID. The hashing based ReID methods can be mainly divided into two categories: \textbf{1)} The method shown in Fig.~\ref{subfig:projections} learns multiple projection matrices to concurrently map original features to a low-dimensional and discriminative Hamming space. However, its objective is generally a non-convex joint function of several sub-tasks (\eg similarity-preserving mapping and binary transformation), which requires the explicit design of sophisticated functions and time-consuming non-convex (discrete) optimizations.
%In addition, traditional projection learning methods are based on matrix operations, which need to deal with all the training data at once.
The memory storage and computational efficiency are serious issues, especially when dealing with large-scale data. \textbf{2)} Fig.~\ref{subfig:deephashing} shows a deep neural network based method, which is able to process large-scale data much more efficiently compared with traditional methods by using mini-batch learning algorithms and advanced GPUs. The binary codes here are generated by inserting hashing layers at the end of the networks. However, the hashing layer is simply a fully-connected layer followed by a tanh-like activation to force the outputs in binary form. This straightforward scheme hardly constrains the outputs under the important principles of hashing (\eg balancedness and independence \cite{weiss2009spectral}) to obtain high-quality binary codes. Moreover, the outputs of the hashing layers tend to lie in the approximately linear part of the tanh-like functions for preserving discriminability. Therefore, directly binarizing the outputs by the $sign$ function will lose the discriminative information.

To address the above issues, this paper proposes an unified end-to-end deep learning framework for efficient ReID, aiming to jointly learn a discriminative feature representation, an accurate similarity measurement and an \emph{implicit} binary transformation. In particular, we propose Adversarial Binary Coding (ABC) by adopting a \emph{Generative Adversarial Net} (GAN) \cite{goodfellow2014generative,radford2015unsupervised} to regularize features into binary form without loss of discriminability (see Fig.~\ref{fig:ABC}). Instead of explicit projections, the adversarial learning makes the target distribution (in binary form) an \emph{`expert'} that implicitly guides the network to generate samples under the same distribution. Specifically, we employ the Bernoulli distribution to guide a CNN to generate discrete features. Benefiting from the nature of the Bernoulli distribution, our ABC can generate high-quality discriminative codes complying with the important principle of hashing, \eg balancedness. As shown in Fig.~\ref{subfig:abc}, our strategy avoids both time-consuming explicit projection learning and low-quality codes with the simple tanh-like activation. More importantly, our ABC can be flexibly embedded into any similarity regressive networks (\eg deep triplet networks) and optimized jointly with the network in an end-to-end manner. The main contributions of this paper are summarized as follows:

1) We propose a binary transformation strategy based on deep adversarial learning. The proposed architecture is composed of a CNN for feature extraction and a discriminator network for distinguishing real-valued and binary features, where the CNN is guided to generate features in binary form to confuse the discriminator. Thus, the features are implicitly regularized into binary codes.

2) An end-to-end deep neural network that seamlessly accommodates the above adversarial binary coding module is built for efficient ReID. We jointly optimize the binary transformation and similarity measurement. Consequently, the discriminative information is largely preserved during feature binarization.

3) Extensive experiments on three large-scale ReID benchmarks (\ie CUHK03 \cite{li2014deepreid}, Market-1501 \cite{zheng2015scalable}, and DukeMTMC-reID \cite{zheng2017unlabeled}) clearly demonstrate the superiority of our framework both in terms of accuracy and efficiency, compared with other binary coding based and the state-of-the-art ReID methods.

%-------------------------------------------------------------------------
\section{Related work}
\label{sec:related}

% In this section, we review the methods of person re-identification in three aspects, namely feature representations, similarity measuring, and deep learning based frameworks. Subsequently, we introduce some key developments of GANs related to our method.

\paragraph{Person re-identification:}
\label{subsec:relatedreid}
% robust low-level appearance representations
% color: hirzer2012relaxed, gray2008viewpoint,
% texture: gray2008viewpoint,
% matching: lisanti2014matching, chen2015relevance, wang2016person
% to map feature vectors to the metric spaces in which the intra-distances are smaller and inter-distances are larger
Traditional approaches usually propose certain feature learning algorithms for ReID, including low-level color features \cite{farenzena2010person,koestinger2012large,pedagadi2013local} and local gradients \cite{prosser2010person,lisanti2015person,liu2015spatio}, and high-level features \cite{liao2015person,lan2016quaternionic,Matsukawa_2016_CVPR}. Due to the breaking-through performance of deep neural networks, deep learning based ReID methods \cite{li2014deepreid,ahmed2015improved,Xiao_2016_CVPR,ZhouSanping_2017_CVPR,Lin_2017_CVPR,Panda_2017_CVPR,Chen_2017_CVPR} have been proposed increasingly. For instance, siamese CNNs \cite{yi2014deep,shi2016embedding,varior2016gated} and triplet CNNs \cite{chen2016deep,cheng2016person,wang2016joint} are widely used for similarity measurement. Very recently, several binary coding based approaches \cite{zhao2015deep,zhang2015bit,zheng2016learning,ChenJiaxin_2017_CVPR} have emerged to deal with the high computation and storage costs existing in ReID problems.

\paragraph{Generative adversarial nets:}
\label{subsec:relatedgan}
GANs \cite{goodfellow2014generative,radford2015unsupervised} provide a methodology to map random variables from a simple distribution to a certain complex one, and have been widely used in image generating \cite{radford2015unsupervised,huang2017stacked,huang2017beyond,mao2017least}, style transferring \cite{zhu2017unpaired,pix2pix_2017_CVPR} and latent feature learning \cite{makhzani2015adversarial,donahue2016adversarial,dumoulin2016adversarially}. To stabilize and quantify the training of GANs, a breakthrough named Wasserstein GAN (WGAN) was proposed in \cite{arjovsky2017wasserstein} and improved in \cite{gulrajani2017improved}. More recently, GANs are also utilized for image retrieval problems. In \cite{qiu2017deep}, GANs were adopted to distinguish synthetic and real images, aiming to improve the discriminability of binary codes. GANs were also employed to enhance the intermediate representation of the generator in \cite{song2018ganbinary}. However, these kinds of studies still simply adopted tanh-like activations for binarization. To our best knowledge, our ABC is the first work that intuitively adopts the spirit of adversarial learning to perform binary transformation for efficient ReID.

%-------------------------------------------------------------------------
\section{Approach}
\label{sec:method}

The proposed framework transforms high-dimensional real-valued features to compact binary codes mainly based on the adversarial learning. In the following, we first briefly review the principles of GANs in Section \ref{subsec:reviewgan}. Then, we introduce the adversarial binary coding (ABC) in detail in Section \ref{subsec:dalbc}. In Section \ref{subsec:triplet}, we present the joint end-to-end framework with triplet networks for efficient ReID.

\subsection{A Brief Review of GANs}
\label{subsec:reviewgan}

In the framework of GAN, there is a generator $G$ which competes against an adversary, \ie a discriminative model $D$ that learns to determine whether a sample is from the model distribution or the data distribution. To learn the generator's distribution $p_g$ over data $x$, a prior on input noise variables is defined as $p_z(z)$, then a mapping to data space is denoted as $G(z; \theta_g)$, where $G$ is a differentiable function represented by a deep neural network with parameters $\theta_g$. Meanwhile, $D(x; \theta_d)$ represents the probability that $x$ comes from the data rather than $p_g$. $D$ is trained to maximize the probability of assigning the correct label to both samples from real data and samples from $G$. Simultaneously, $G$ is trained to make $D$ confused. The formal loss function is defined as follows:
\begin{equation}
\begin{split}
\min \limits_{G} \max \limits_{D} V(D, G)& =\mathbb{E}_{x\backsim P_{data}(x)}[\log D(x)]\\
 &+ \mathbb{E}_{z\backsim P_{z}(z)}[\log (1 - D(G(z)))].
\end{split}
%%\vspace{-0.3em}
\label{equ:gan}
\end{equation}

%The method has been successfully verified on several image datasets (\emph{e.g.}\@\xspace MNIST \cite{lecun1998gradient}, CIFAR10 \cite{krizhevsky2009learning}).
However, GANs are difficult to train so that the generator may fail to generate either real-looking or diverse samples. Arjovsky \etal \cite{arjovsky2017towards,arjovsky2017wasserstein} addressed this problem by introducing the WGAN, which optimizes the Wasserstein loss instead of the Jensen-Shannon divergence to evaluate the similarity. The Wasserstein loss is defined as the following:
\begin{equation}
\max \limits_{w} \mathbb{E}_{x\backsim P_{data}(x)}[D(x)] - \mathbb{E}_{z\backsim P_{z}(z)}[D(G(z))].
%%\vspace{-0.4em}
\label{equ:wgan}
\end{equation}
It provides stronger stability of gradients based on the Wasserstein-1 distance (also called the Earth-Mover distance). Moreover, WGAN provides meaningful learning curves useful for debugging and hyper-parameter searching. Therefore, in this work, we adopt the training strategy of WGAN for adversarial learning.

\begin{figure}[t]
\begin{center}
   \includegraphics[width=0.85\linewidth]{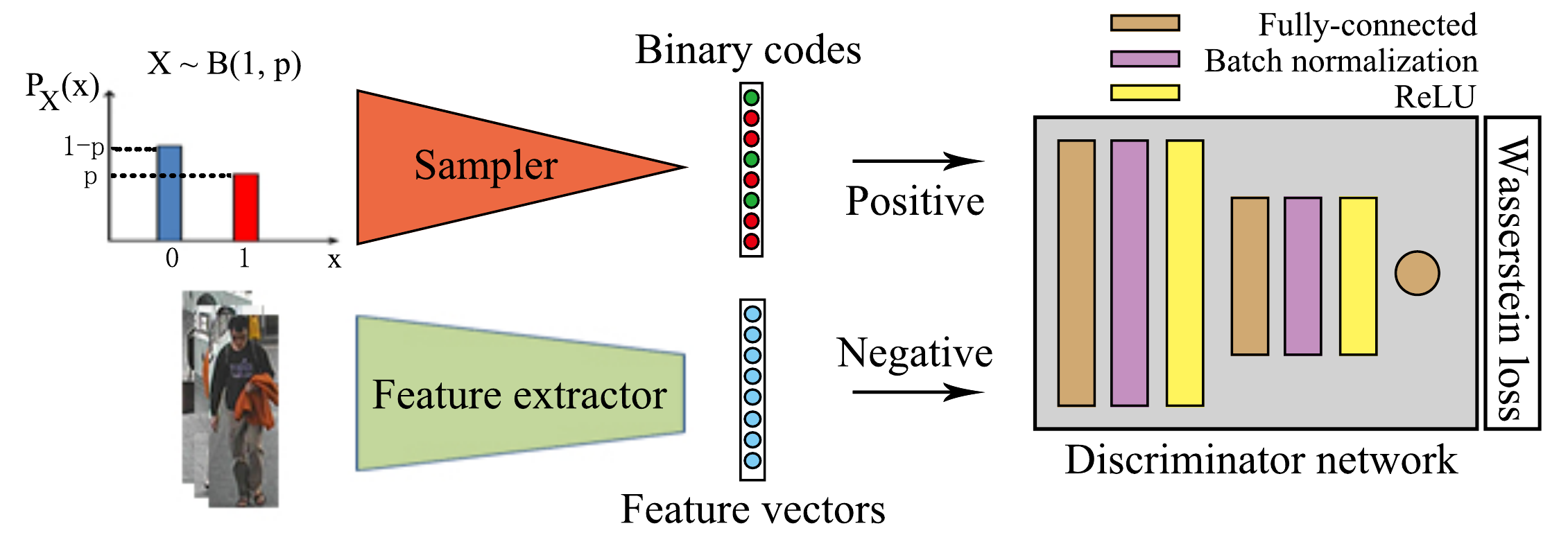}
\end{center}
\vspace{-3.6mm}
\caption{Illustration of the adversarial binary coding (ABC) framework. The discriminator network receives sampled binary codes as positive samples and extracted real-valued features as negative samples. The feature extractor network and discriminator network are jointly optimized under the Wasserstein loss (W loss), such that the extractor is forced to generate features in binary form.}
\label{fig:ABC}
\vspace{2mm}
\end{figure}

\subsection{Adversarial Binary Coding}
\label{subsec:dalbc}

% existing methods need the tedious search of parameters.
% without the tedious search of parameters.
%As mentioned earlier, binary codes have advantages in the efficiency related aspects. Many existing methods attempt to formulate hashing functions in order to achieve explicit binary transformations. However, a designed function maybe unable to keep the semantic information of features and transform them to binary form concurrently for varied data distributions.
Our binary coding scheme is intuitively inspired by GANs. Instead of formulating explicit hashing functions (\ie learning explicit projections), we implicitly guide a deep neural network to directly learn the transformation of data from the original distribution (\ie images) to a distribution of binary vectors in a GAN framework. In this section, we focus on the binary transformation module of the end-to-end efficient ReID framework. How to keep the semantic/discriminative information during the transformation procedure is explained in Section \ref{subsec:triplet}.

%Instead of random noise variables, we transform data under a certain distribution to the target distribution. The first work that proposed similar model is \cite{makhzani2015adversarial}. It adopted a multi-layer perceptrons based encoder-decoder structure and a GAN framework to regularize the intermediate outputs of the network under the Gaussian distribution. Motivated by its success, we attempt to transform data to the discrete binary codes.

The proposed framework of adversarial binary coding is illustrated in Fig. \ref{fig:ABC}. The feature extractor can be any CNN architecture (ResNet-50 \cite{he2016deep} is adopted in this work), which finally represents the images as feature vectors. Meanwhile, a binary code sampler performs random sampling for every bit of the binary vectors. To satisfy the principle of effective binary coding mentioned in \cite{weiss2009spectral}, we sample from the Bernoulli distribution, based on which there is a 50\% chance for each bit to be 0 or 1, and different bits are independent of each other. The discriminator is expected to classify the binary vectors as positive samples and the real-valued feature vectors as negative samples. Thus, the extractor is trained to generate feature vectors that are under the same distribution of positive samples using the Wasserstein loss (W loss) in Eq. (\ref{equ:wgan}).

Formally, we denote a batch of $n$ images as $\textbf{I} = \{\textbf{I}_1, \textbf{I}_2, \dots, \textbf{I}_n\}$ under the distribution $p_{\textbf{I}}$. The feature extractor is denoted as a mapping function $f(\textbf{I})$ which plays the role of the generator $G$ in the definition of GANs under an encoding distribution $q(\textbf{Z}\mid \textbf{I})$ where $\textbf{Z} = \{\textbf{z}_1, \textbf{z}_2, \dots, \textbf{z}_n\}$ denotes the extracted feature vectors. $q(\textbf{Z}\mid \textbf{I})$ aims to transform data from the original distribution $p_\textbf{I}$ to a target distribution $q$:
%%\vspace{-1em}
\begin{equation}
q(\textbf{z}_i) = \int_{\textbf{I}_i} q(\textbf{z}_i\mid \textbf{I}_i)p_\textbf{I}(\textbf{I}_i)d\textbf{I}_i, ~i=1,\dots,n.
%\vspace{-0.5em}
\label{equ:qz}
\end{equation}
Since a binomial distribution is equivalent to multiple Bernoulli samplings with the same probability, the extractor is essentially regularized by matching the posterior $q$ to a prior binomial distribution using the Wasserstein distance.

As mentioned above, we use ResNet-50 \cite{he2016deep} as the backbone model, where Rectified Linear Unit (ReLU) is adopted as the activation function. Hence, we represent every bit of the binary codes by $\{0, 1\}$ instead of $\{-1, 1\}$ \cite{zhang2015bit,ChenJiaxin_2017_CVPR} due to the non-negative outputs of ReLU. We further find that the performance will severely deteriorate if the feature vectors and binary codes are directly fed into the discriminator and the similarity regressive loss (\eg triplet loss in Fig. \ref{fig:framework}) \emph{without} normalization, due to the contradiction between the expected 0 or 1 outputs and the learning algorithm. More concretely, the weights of the neural network are generally initialized to very small values (much smaller than 1). Meanwhile, the learning algorithm carefully controls the scale of the weights (\eg by learning rates and weight decay mechanism) to avoid gradient vanishing or exploding under the loss function. As a result, the features extracted by the network will also be very close to 0 since they share the same scale with weights. On the contrary, our ABC expects every dimension of the output features to be constrained near 0 or 1. As a consequence, we will encounter an unstable optimization process if not adopting any normalization.
% In addition, both the output feature vectors and sampled binary codes are normalized to the same scale by $\ell_2$ normalization.

To address the above issue, we normalize both the output feature vectors and sampled binary codes to the same scale by $\ell_2$ normalization. As for the real-valued features, we adopt the standard $\ell_2$-Norm operation. In terms of the binary codes, we perform the normalization specifically as follows. Given a batch of random binary vectors $ \{ \textbf{B}_{i} \}_{i=1}^{n}$, where $ \textbf{B}_{i} \in \{0,1\}^m$ and $m$ is the code length, the binary vectors can be directly normalized as follows:
\begin{equation}
\widetilde{\textbf{B}_{i}} = \frac{\textbf{B}_i}{\| \textbf{B}_i\|_2}, ~i=1,\dots,n.
%\vspace{-0.3em}
\label{equ:naivenorm}
\end{equation}

\noindent{}However, the $\ell_2$-Norm of every vector could be different since every binary vector may contain different numbers of bits assigned to $1$. In other words, the values of non-zero entries in the normalized vectors $\{ \widetilde{\textbf{B}_{i}} \}_{i=1}^n$ will be different. This leads to an unstable training process, where the losses are unable to guide the optimization clearly. Therefore, in this study, we adopt the expectation of the Bernoulli distribution to calculate the $\ell_2$-Norm of binary vectors. Specifically, we calculate an uniform normalization factor $\lambda$ as:
\begin{equation}
\lambda = (\sum_{j=1}^{m}\mathbb{E}_{Bernoulli}[1]^2)^{\frac{1}{2}},
\label{equ:normlambda}
\end{equation}
where $\mathbb{E}_{Bernoulli}[1]$ represents the expectation of Bernoulli random variables, and thus the binary vectors can be normalized as $\widetilde{\textbf{B}_{i}} = \frac{1}{\lambda} \textbf{B}_{i}, ~i=1,\dots,n$.

%-------------------------------------------------------------------------
\subsection{Triplet Loss based Efficient ReID Framework}
\label{subsec:triplet}

\begin{figure}[t]
\begin{center}
   \includegraphics[width=0.9\linewidth]{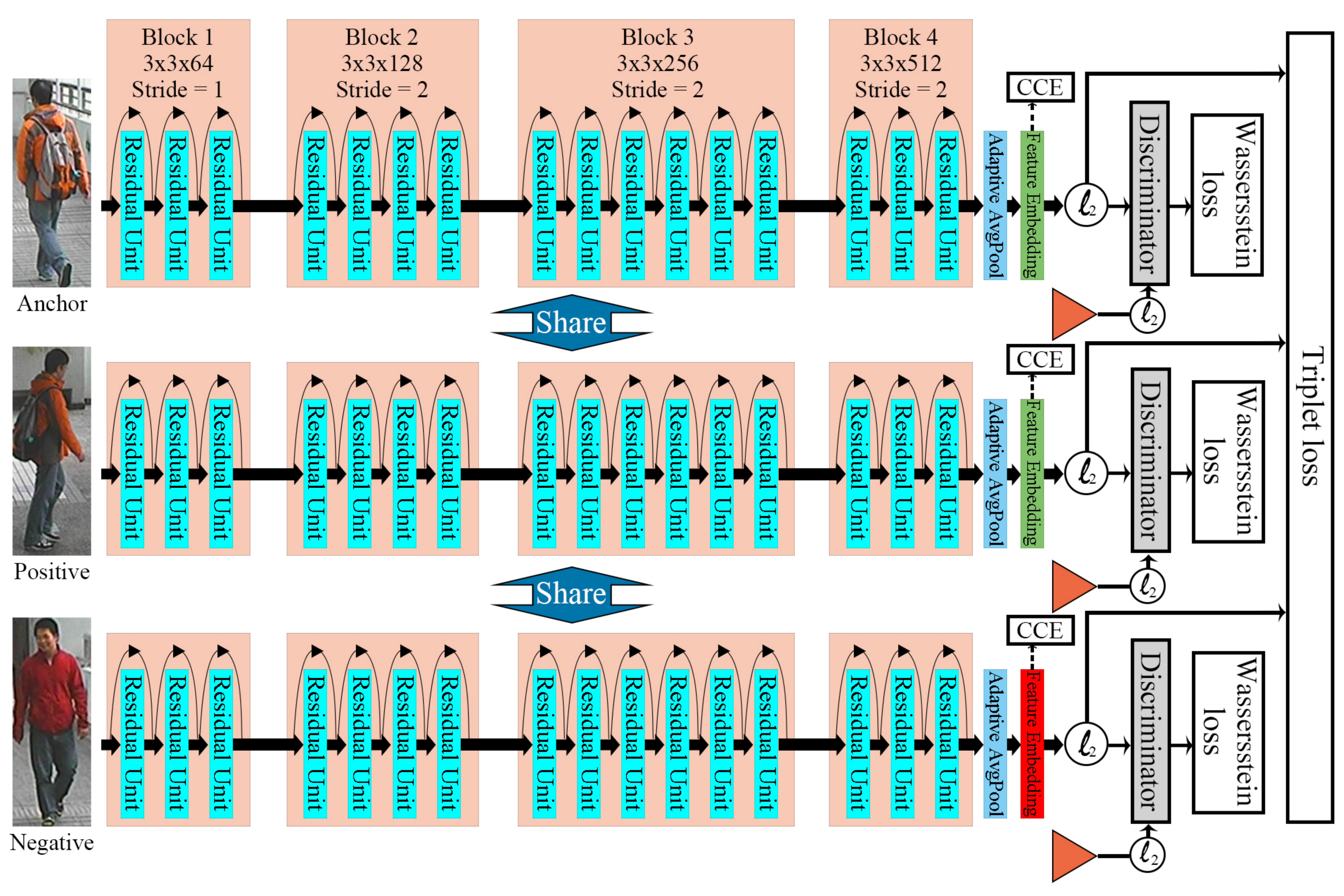}
\end{center}
\vspace{-3.6mm}
   \caption{Illustration of the triplet loss based adversarial binary coding embedded efficient ReID framework. The feature extraction network is a pre-trained ResNet-50 model. The network is first fine-tuned by the Cross-Entropy Error (CCE) loss on training images and then simultaneously trained by the triplet loss and the Wasserstein loss to generate discriminative binary features.}
   \label{fig:framework}
\vspace{1.5mm}
\end{figure}

To not only transform features to binary form, but also measure similarities between binary codes, the ABC is further embedded into a triplet network for ensuring the discriminability of the learned binary codes. The triplet loss \cite{balntas2016learning} $L$ is formulated as follows:
%%\vspace{-1.1em}
\begin{equation}
L(x_i, x_j, y_k, \alpha) = \frac{1}{n} \sum_{i=1}^{n}\max \{ d(x_i,x_j) - d(x_i,y_k) + \alpha , 0 \},
%%\vspace{-0.3em}
\label{equ:triplet}
\end{equation}
where $x_i$, $x_j$, and $y_k$ are input features, $\alpha$ is the imposed distance margin between positive and negative pairs, and $d(\cdot)$ measures the similarity distance. $x_i$ and $x_j$ are features from the same class (same identity in ReID), and $y_k$ is from another class (different identity). The triplet loss forces the distances between samples in negative pairs larger than those in positive pairs. Therefore, it is widely used in the tasks which aim to retrieve data with high relevance.

The overall framework for efficient ReID is shown in Fig. \ref{fig:framework}. ResNet-50 pre-trained on ImageNet \cite{imagenet2009} is adopted as the backbone model, where the fixed average pooling layer is replaced by an adaptive average pooling to fit different input sizes, followed by a feature embedding (fully-connected) layer to reduce feature dimensions into expected lengths. At the beginning of training, we fine-tune the model on pedestrian images with Cross Entropy Error (CEE) loss, by solving a conventional classification problem, \ie each class contains the images of one person. Because fine-tuning the models pre-trained on a large image collection on small datasets has been verified as an effective approach for knowledge transfer. This is also helpful for deep networks with less data to find the optimal parameters more easily and to reach the convergence faster compared with training from scratch. Note that the outputs of the last layer are not normalized by $\ell_2$-norm in this phase, just as conventional CNNs for image classification. After that, we train the model with normalization as shown in Fig. \ref{fig:framework}, by jointly optimizing the Wasserstein loss for binary coding and triplet loss for similarity measuring. Specifically for the composition of triplet batches, we randomly select $n$ different persons and pick two images from different views of each person to be the anchor and positive samples. Then we randomly select an image of a person different from the anchor as the negative sample in each triplet.

%%% TODO: Hard Positive and Negative Mining??
% More specifically for the triplet selection, we repeatedly compose a triplet batch in the following steps when using triplet loss. (1) Randomly select different persons, the number of which is same to the batch size, and use one of the images of every person from one viewpoint as the anchors in the triplets. (2) Randomly select another viewpoints and an image from it of each corresponding person as the positive samples. (3) Randomly select persons different from the corresponding anchor persons and randomly select images of them to be the negative samples. When training the model in the DAL-BC framework, the samples in a batch are randomly picked from all images in the training set.

% ... better derivation thus better gradients, same results.
Particularly, in the training phase, we adopt the Euclidean distance to measure similarities between real-valued features for the triplet loss without binarization. Because the Euclidean distance provides conspicuously more stable gradients than the Hamming distance, while obtaining the equivalent distance measuring results as the Hamming distance. In this way, the triplet loss focuses on reducing intra-class distances and enlarging inter-class distances in terms of the real-valued features, whilst the Wasserstein loss focuses on the binary transformation of real-valued features.

% the distance to measure similarities in the triplet loss function is still the Euclidean distance instead of the Hamming distance, since the binarization is not executed during training. This is because that the Euclidean distance gives same results to the Hamming distance since we choose $\{0, 1\}$ as codes. Although computing Euclidean distances is notably slower than computing Hamming distances, it provides much stable

In the testing phase, images are sent into the trained CNN to obtain the real-valued features, of which every entry should be very close to binary values, such as $\epsilon$ or $\lambda - \epsilon$, where $\epsilon$ is a very small value. Finally, we binarize the features as follows:
%%\vspace{-0.8em}
\begin{equation}
b_j =
\begin{cases}
  0& z_j \le \frac{1}{2\lambda};\\
  1& z_j > \frac{1}{2\lambda},
\end{cases}
 ~j=1,\dots,m,
\label{equ:threshold}
\end{equation}
where $z_j$ is the value of the $j$-th entry of a real-valued feature $\textbf{z}_i = [z_1, \dots, z_m]\in\mathbb{R}^m$ extracted by $f(\textbf{I}_i)$, and $b_j\in\{0,1\}$ is the binary bit of $z_j$ after binarization. The Hamming distances between queries and the gallery set are further computed using extremely fast XOR operations to measure similarities.

%-------------------------------------------------------------------------
\section{Experiments}
\label{sec:experiments}

We evaluate the performance of our method on three large-scale ReID datasets: CUHK03 \cite{li2014deepreid}, Market-1501 \cite{zheng2015scalable}, and DukeMTMC-reID \cite{ristani2016MTMC,zheng2017unlabeled}. The goal of our experiments is mainly to answer the following three research questions:
\begin{itemize}
	\item \textbf{Q1:} How efficiency in computation and storage is our learned binary codes, compared with their real-valued counterparts? (Sec. \ref{subsec:efficiency})
	\item \textbf{Q2:} How does our method perform compared to the binary coding based ReID approaches (Sec. \ref{subsec:cmphash}), and the state-of-the-art non-hashing ReID approaches (Sec. \ref{subsec:cmpstateoftheart})?
	\item \textbf{Q3:} How will our performance change with different configurations (\eg different similarity networks, with and without $\ell_2$-norm/fine-tuning)? (Sec. \ref{subsec:ablation})
\end{itemize}

\subsection{Datasets and Settings}
\label{subsec:dataset}

%CUHK01 \cite{li2012human},
\textbf{CUHK03} contains 14,096 images of 1,467 identities captured by six surveillance cameras. The dataset provides both manually labeled and automatically detected bounding boxes with variant sizes. In the experiments, we resize the images to 160$\times$60, and the 20 training/testing splits reported in \cite{li2014deepreid} are used. The number of training iterations is set to 6,000, and the margin of the triplet loss is initialized to 0.2 and increased to 0.3 after 1,000 iterations, 0.4 after 2,500 iterations, and 0.5 after 4,000 iterations.

\noindent\textbf{Market-1501} contains 32,668 automatically detected 128$\times$64 bounding boxes of 1,501 pedestrians under six cameras and provides a fixed evaluation protocol. In the experiments, the number of training iterations is set to 8,000, and the margin of the triplet loss is initialized to 0.2 and increased to 0.3 after 1,000 iterations and 0.4 after 4,000 iterations.

\noindent\textbf{DukeMTMC-reID} contains 36,411 manually annotated bounding boxes of 1,404 identities under 8 cameras and provides a fixed training/testing split. In the experiments, the size of images is resized to 128$\times$64. The number of training iterations is set to 8,000, and the margin of the triplet loss is initialized to 0.2 and increased to 0.3 after 2,000 iterations and 0.4 after 5,000 iterations.

\begin{figure}[t]
\begin{center}
%\subfigure[CUHK01 (486)]{ \label{subfig:cuhk01timememory}
%  %\fbox{\rule{0pt}{1in} \rule{0.21\linewidth}{0pt}}
%  \includegraphics[width=0.3\linewidth]{cuhk01_486_float_binary-eps-converted-to.pdf}
%}
%\hspace{0.01in}
\subfigure[Market-1501]{ \label{subfig:markettimememory}
  %\fbox{\rule{0pt}{1in} \rule{0.21\linewidth}{0pt}}
  \includegraphics[width=0.4\linewidth]{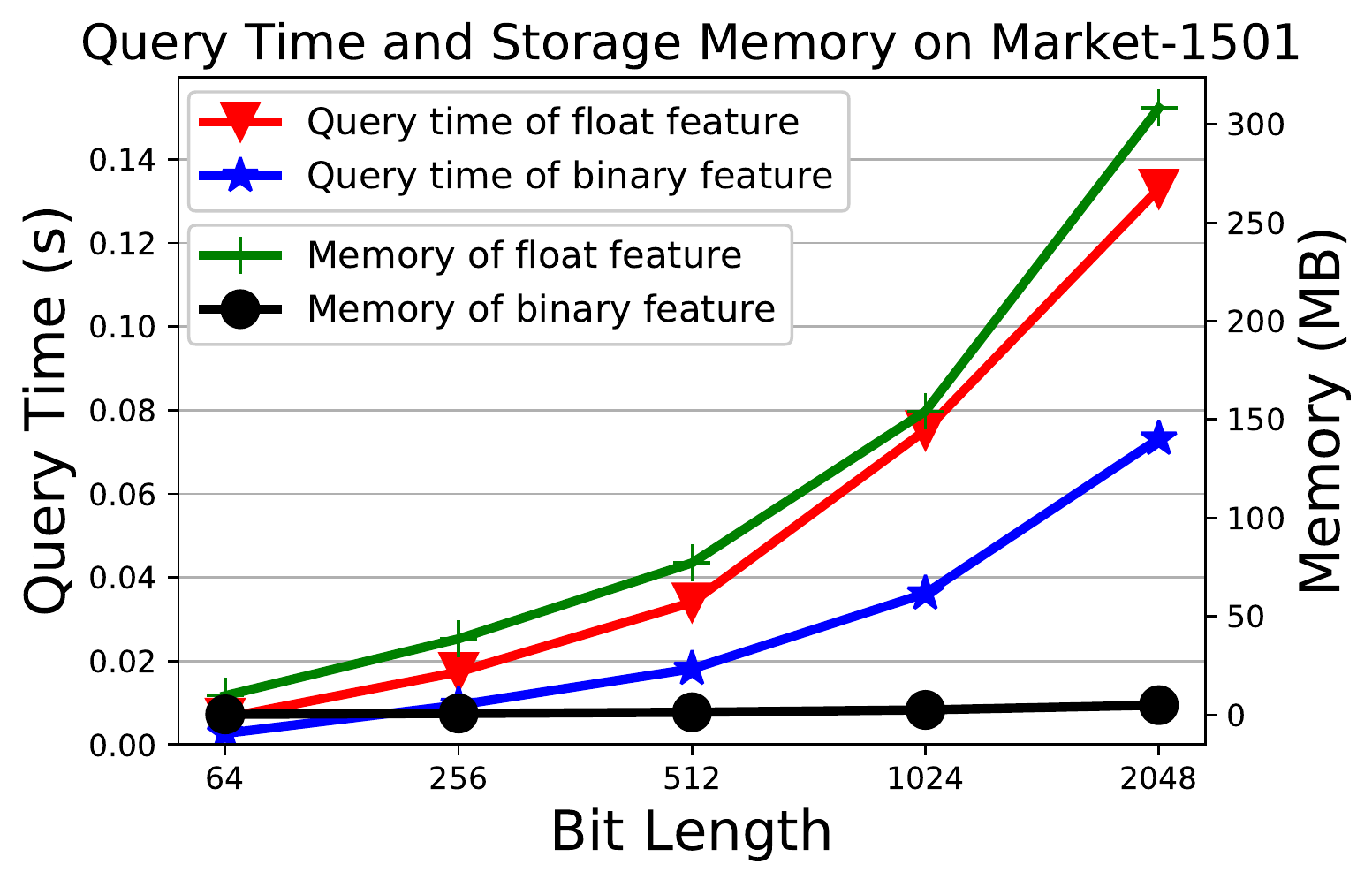}
}
\hspace{6mm}
\subfigure[DukeMTMC-reID]{ \label{subfig:duketimememory}
  %\fbox{\rule{0pt}{1in} \rule{0.21\linewidth}{0pt}}
  \includegraphics[width=0.4\linewidth]{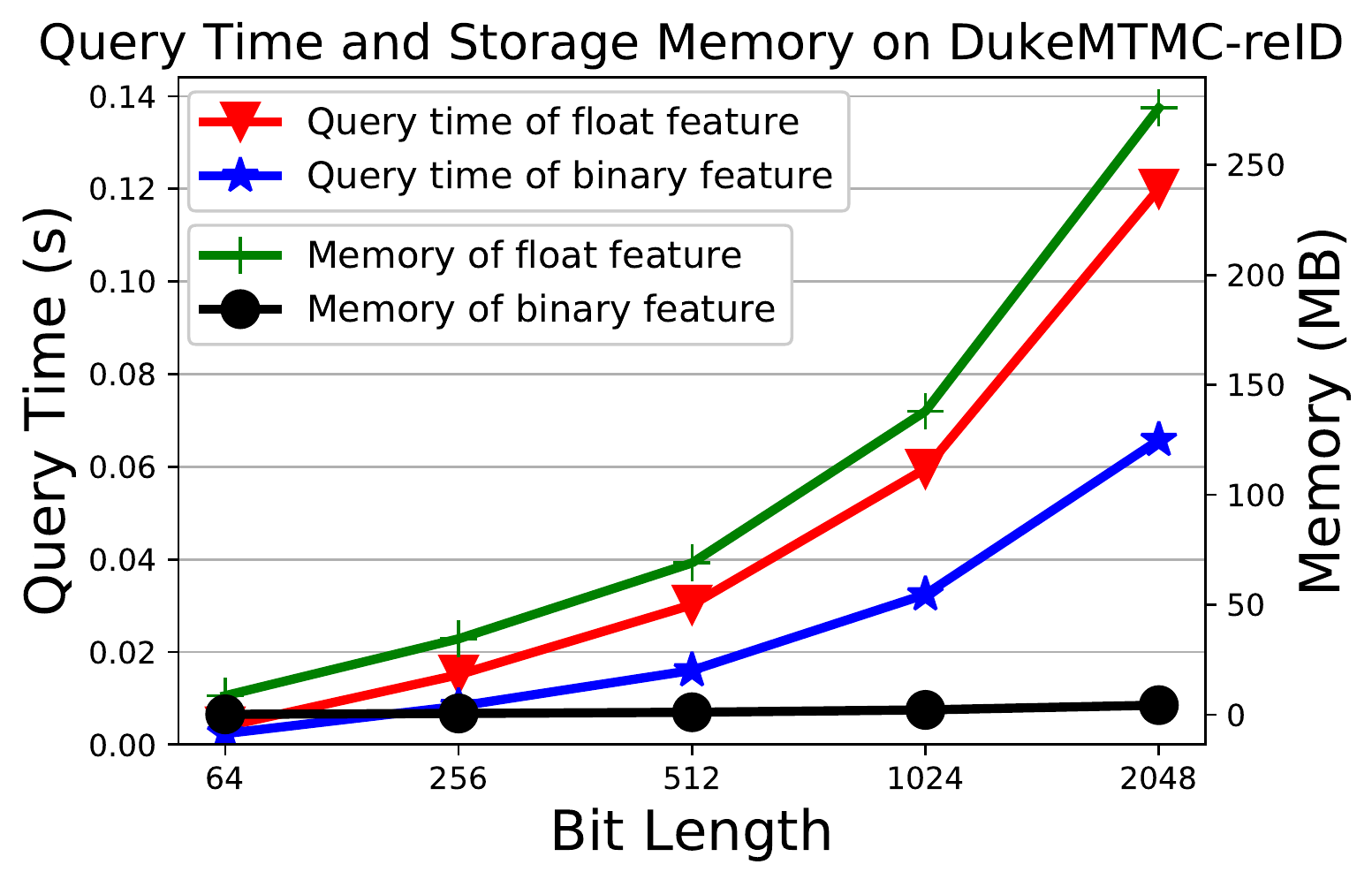}
}
\end{center}
\vspace{-3.6mm}
   \caption{Comparison of matching time and memory costs of real-valued/binary features in terms of different bit lengths.}
\label{fig:timememory}
\vspace{2mm}
\end{figure}

\begin{table}[b]%\Large
\begin{center}
\newcommand{\tabincell}[2]{\begin{tabular}{@{}#1@{}}#2\end{tabular}}
%%\vspace{-0.5em}
\caption{Rank-1 matching rates (\%) of real-valued and binary features on different datasets by using the proposed ABC embedded triplet network.}
\label{table:floatbinary}
%\vspace{-2mm}
\resizebox{.98\textwidth}{!}{
\begin{tabular}{r|cc|cc|cc|cc}
\hline
\multirow{2}{*}{\tabincell{c}{\textbf{Bit Length}}}
& \multicolumn{2}{c|}{ \textbf{CUHK03 (labelled)} }& \multicolumn{2}{c|}{ \textbf{CUHK03 (detected)} } & \multicolumn{2}{c|}{ \textbf{Market-1501} } & \multicolumn{2}{c}{ \textbf{DukeMTMC-reID} } \\
\cline{2-9}
& ~~real-valued~~ & binary & ~~~real-valued~~~ & binary & real-valued & binary & ~real-valued~ & binary\\
\hline
64-bit   & 51.7 & 42.0 & 49.5 & 43.1 & 52.2 & 37.1 & 58.0 & 48.8\\
128-bit  & 55.3 & 46.2 & 54.5 & 48.4 & 57.7 & 44.5 & 67.9 & 60.3\\
256-bit  & 57.6 & 52.9 & 56.8 & 50.3 & 61.6 & 49.6 & 71.4 & 63.5\\
512-bit  & 60.5 & 58.2 & 61.4 & 57.8 & 67.5 & 59.3 & 75.1 & 65.5\\
1024-bit & 61.7 & 60.4 & 65.6 & 61.2 & 70.7 & 66.8 & 76.9 & 69.7\\
2048-bit & 69.4 & 68.8 & 68.9 & 68.1 & 75.8 & 73.5 & 80.3 & 77.6\\
\hline
\end{tabular}
}
\end{center}
%\vspace{-3mm}
\end{table}

We implement our framework based on the PyTorch deep learning library. The hardware environment is a PC with Intel Core CPUs (3.4GHz), 32 GB memory, and an NVIDIA GTX TITAN X GPU. For all the datasets, the images are horizontally flipped to augment training samples. The batch size is set to 64 in the pre-training phase and altered to 128 in the subsequent training. The learning rate of the extractors in the experiments is initialized to 0.001 and decreased to 0.0001 with the iterations. The learning rate of the discriminator is consistently set to 0.01. To ensure the stability, we update the GAN 10 iterations alone  after every 20 global optimization iterations. Every GAN iteration consists of 5 iterations of discriminator updating and 1 iteration of generator (extractor) updating. In the experiments, we re-run the comparison methods if the codes are publicly available to evaluate their efficiencies for fair comparisons.

\subsection{Evaluation of Computation and Storage Efficiency}
\label{subsec:efficiency}

%\begin{table*}[!htb]%\scriptsize
%\begin{center}
%\newcommand{\tabincell}[2]{\begin{tabular}{@{}#1@{}}#2\end{tabular}}
%%%\vspace{0.6em}
%\caption{Rank 1 matching rates (\%) with different bit length using different hashing approaches on CUHK03 (labelled) dataset.}
%\label{tab:cuhk03hashing}
%%%\vspace{-1.7em}
%\resizebox{0.7\textwidth}{!}{
%\begin{tabular}{|c|c|c|c|c|}
%\hline
%\textbf{Method} & \textbf{64-bit} & \textbf{128-bit} & \textbf{256-bit} & \textbf{512-bit}\\
%\hline
%\footnotesize
%SePH \cite{lin2015semantics}          & 1.3  & 1.6  & 1.9  & 2.2  \\
%SCM  \cite{zhang2014large}            &  2.0 & 2.0  & 2.2  & 2.1  \\
%COSDISH \cite{kang2016column}         & 4.4  & 9.3  & 19.1 & 29.0 \\
%SDH \cite{shen2015supervised}         & 12.9 & 19.3 & 25.0 & 31.4 \\
%DRSCH \cite{zhang2015bit}             & 22.0 & 18.7 & -    & -    \\
%DSRH \cite{zhao2015deep}              & 14.4 & 8.1  & -    & -    \\
%FastHash \cite{lin2014fast}           & 2.6  & 4.9  & 8.6  & 12.1 \\
%KSH \cite{liu2012supervised}          & 19.1 & 18.0 & 15.3 & 15.0 \\
%CSBT \cite{ChenJiaxin_2017_CVPR}      & \textcolor{blue}{33.1} & \textcolor{blue}{36.2} & \textcolor{blue}{40.3} & \textcolor{blue}{46.2} \\
%\hline
%ABC+triplet (Ours)                            & \textcolor{red}{42.0} & \textcolor{red}{46.2} & \textcolor{red}{52.9} & \textcolor{red}{58.2} \\
%\hline
%\end{tabular}
%}
%\end{center}
%%\vspace{-5mm}
%\end{table*}

\begin{figure}[t]
\begin{center}

\subfigure[Triplet loss]{ \label{subfig:cuhk03loss_1}
\vspace{-2mm}
  \includegraphics[width=0.3\linewidth]{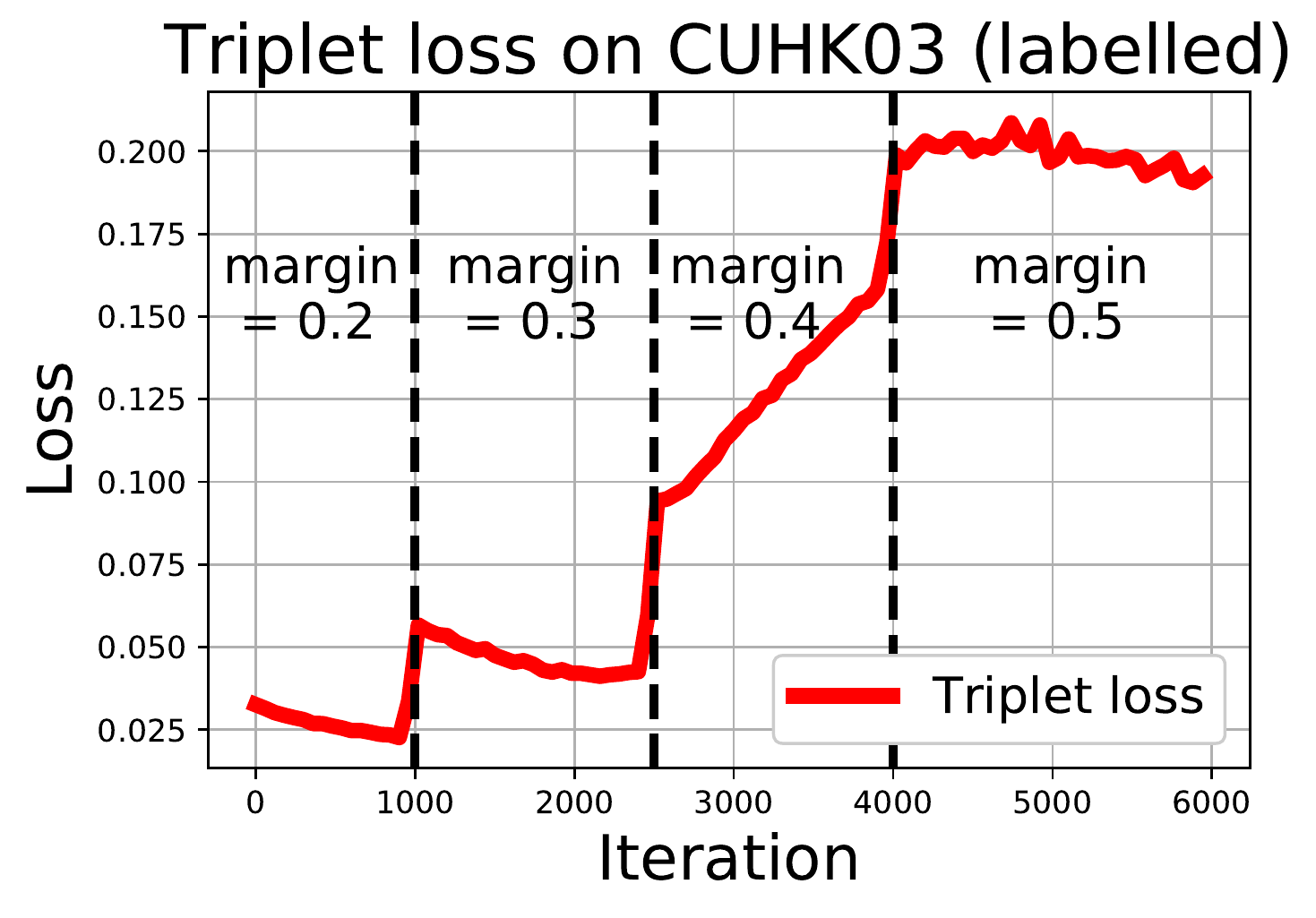}
  %\hspace{-1em}
  \includegraphics[width=0.3\linewidth]{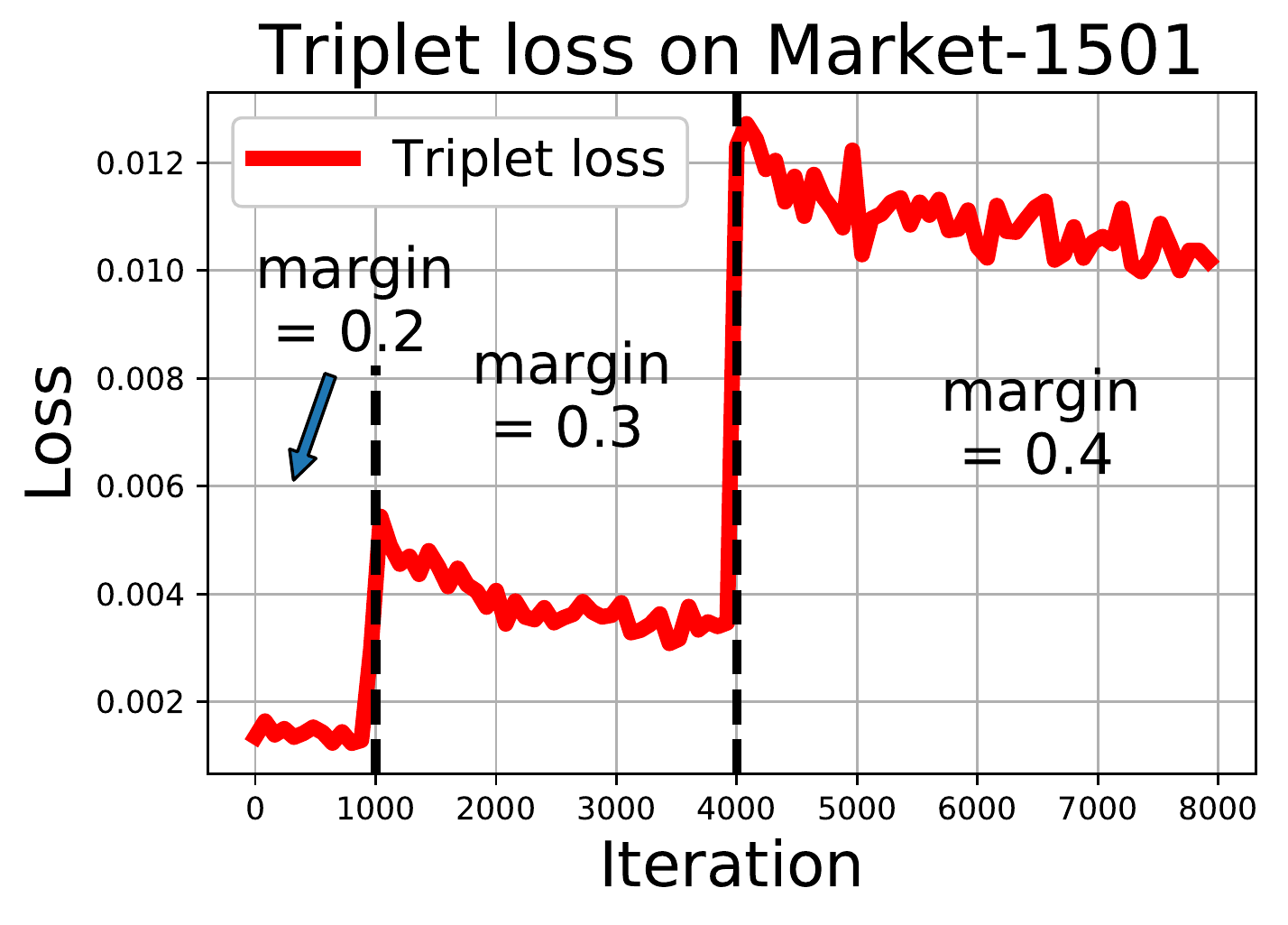}
  %\hspace{-1em}
  \includegraphics[width=0.3\linewidth]{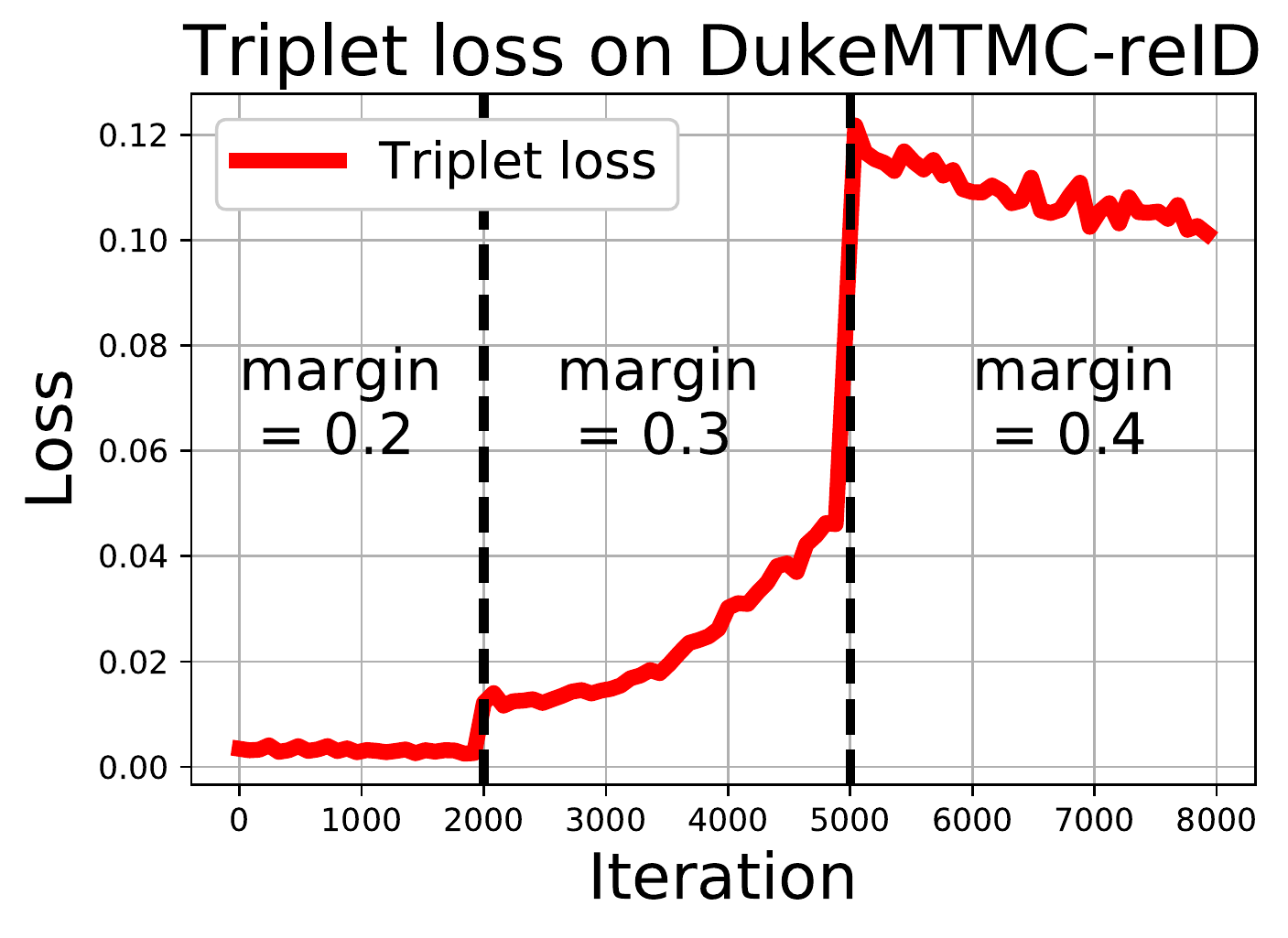}
}
%\vspace{-3mm}
\subfigure[W loss of D]{ \label{subfig:marketloss_1}
  \includegraphics[width=0.3\linewidth]{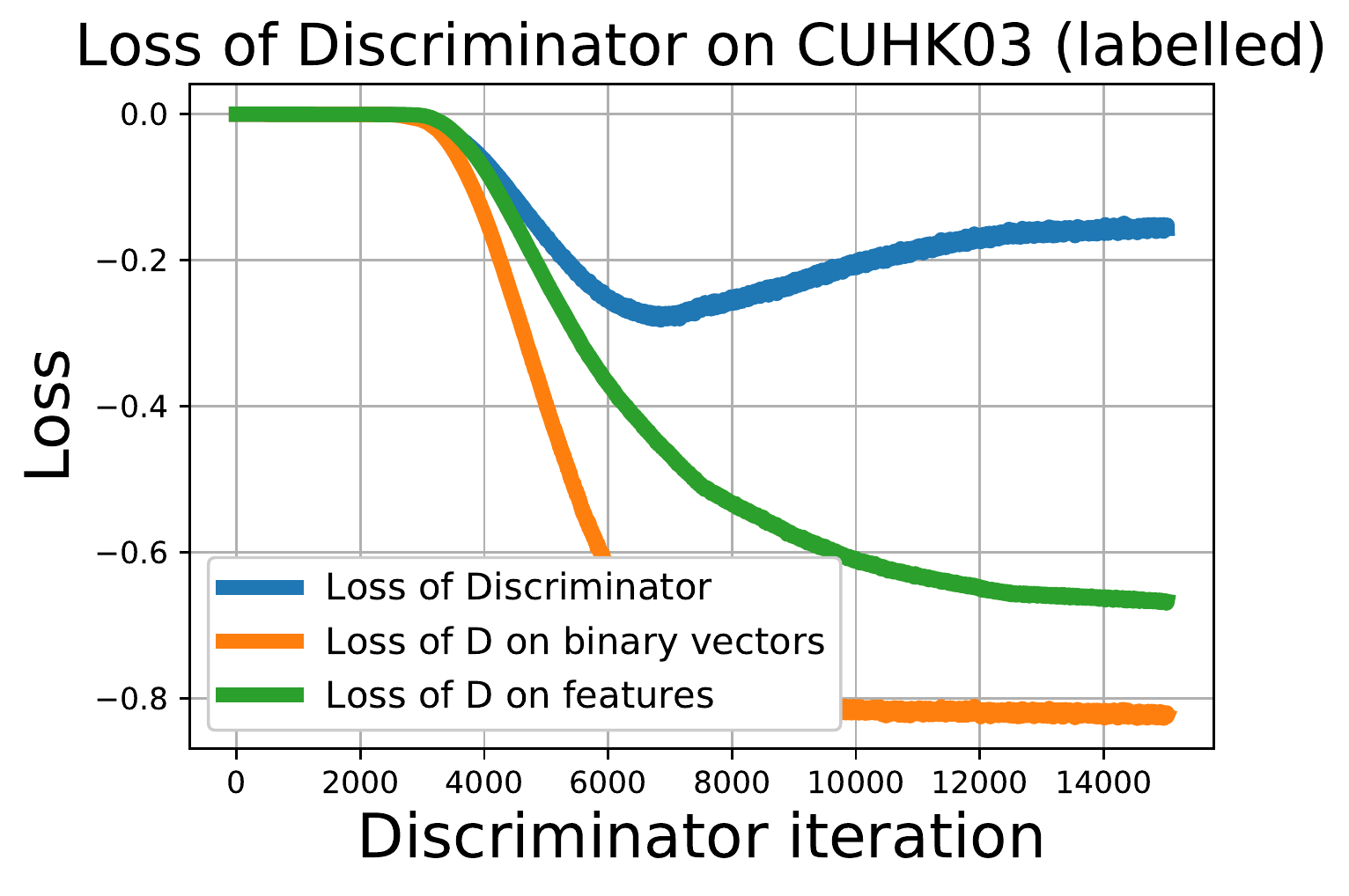}
  %\hspace{-1em}
  \includegraphics[width=0.3\linewidth]{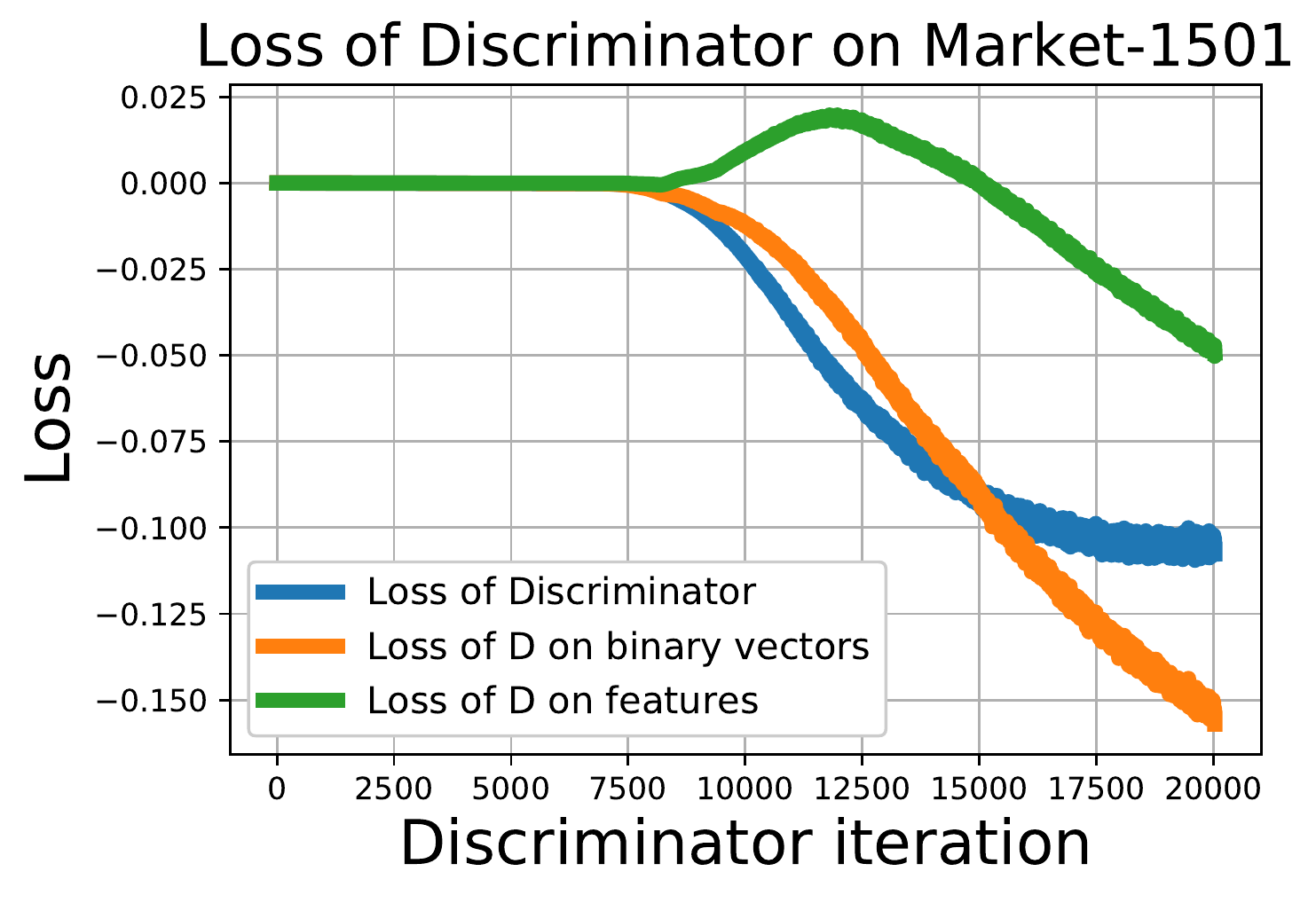}
  %\hspace{-1em}
  \includegraphics[width=0.3\linewidth]{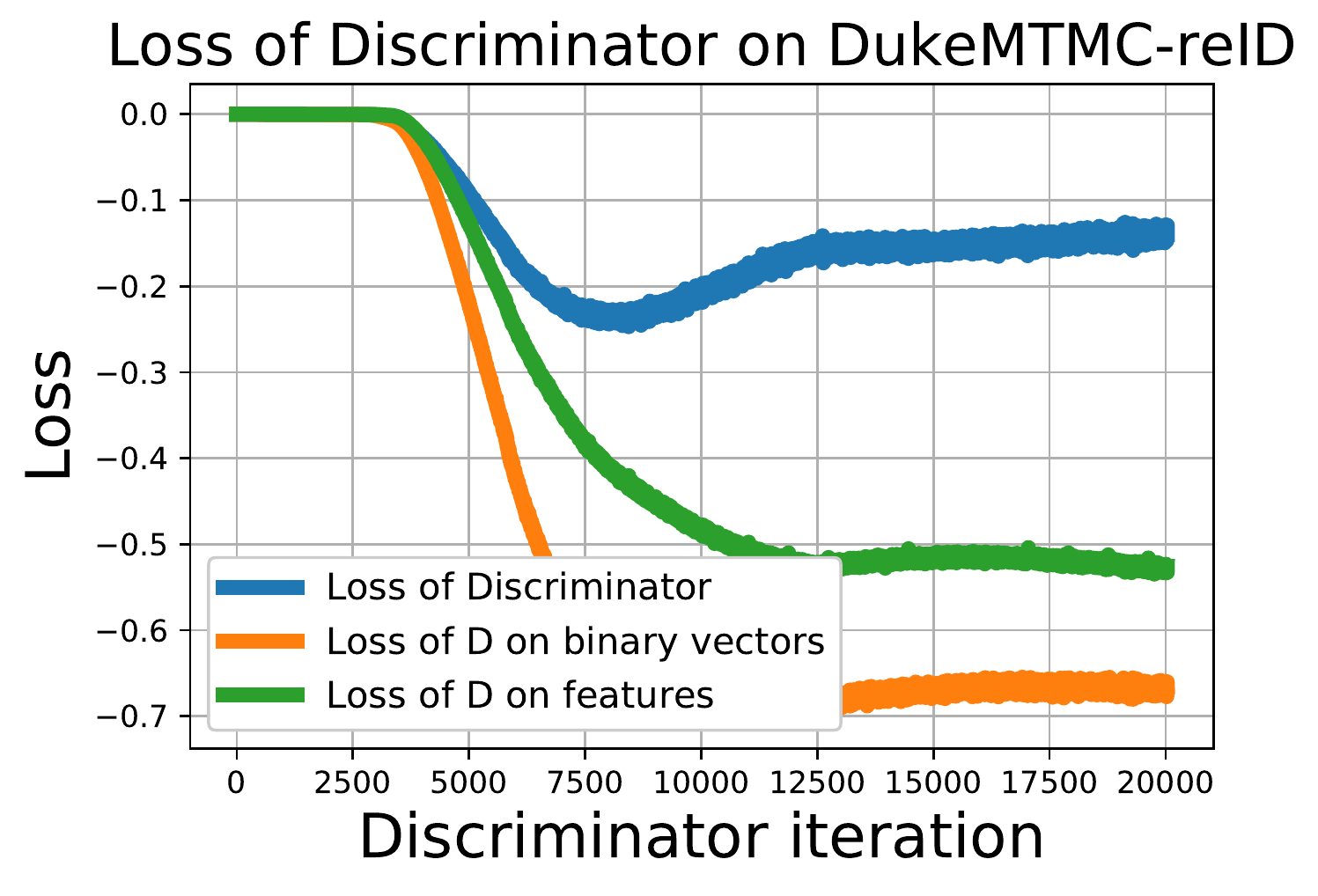}
}

\subfigure[W loss of G]{ \label{subfig:dukeloss_1}
  \includegraphics[width=0.3\linewidth]{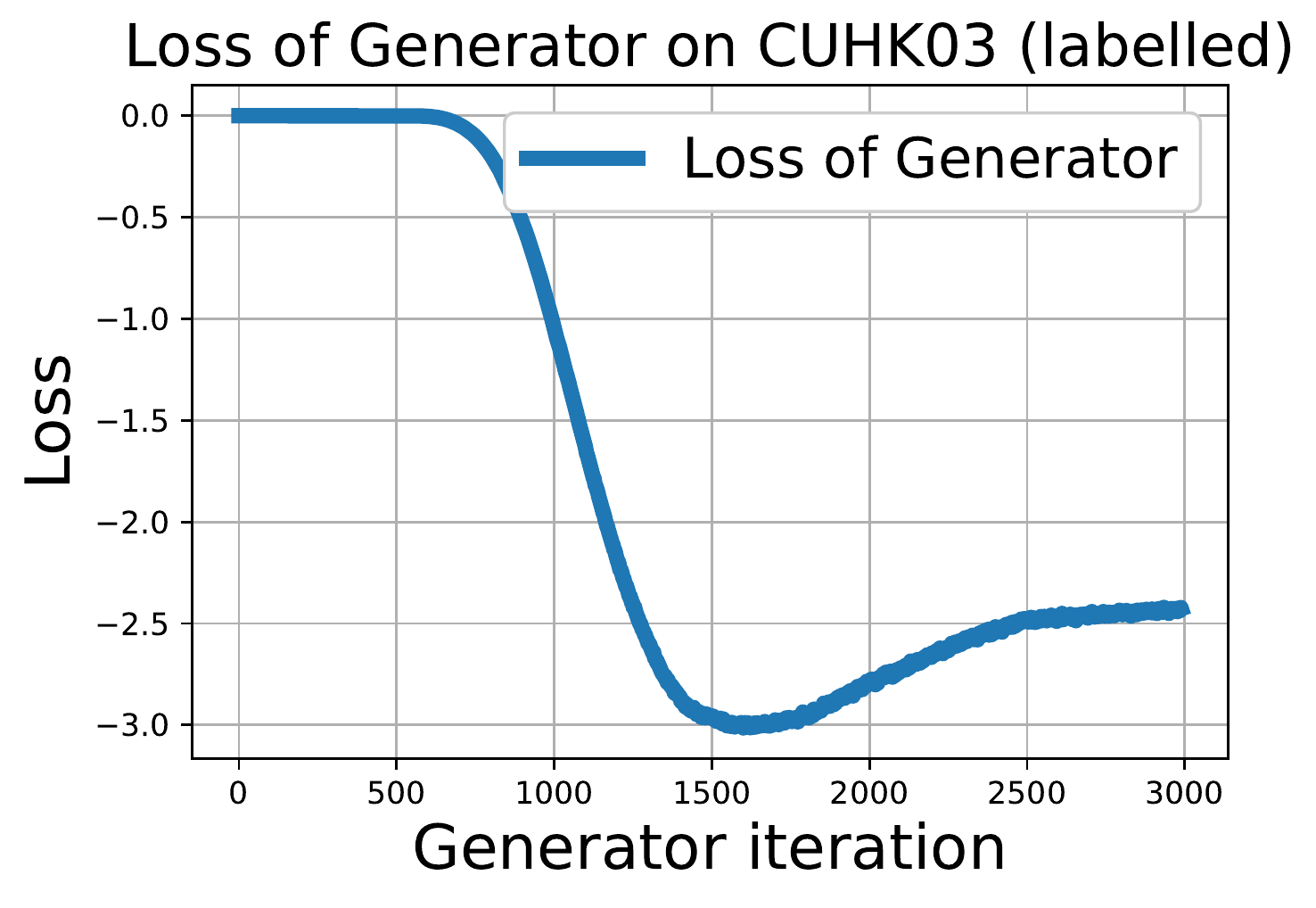}
  %\hspace{-1em}
  \includegraphics[width=0.3\linewidth]{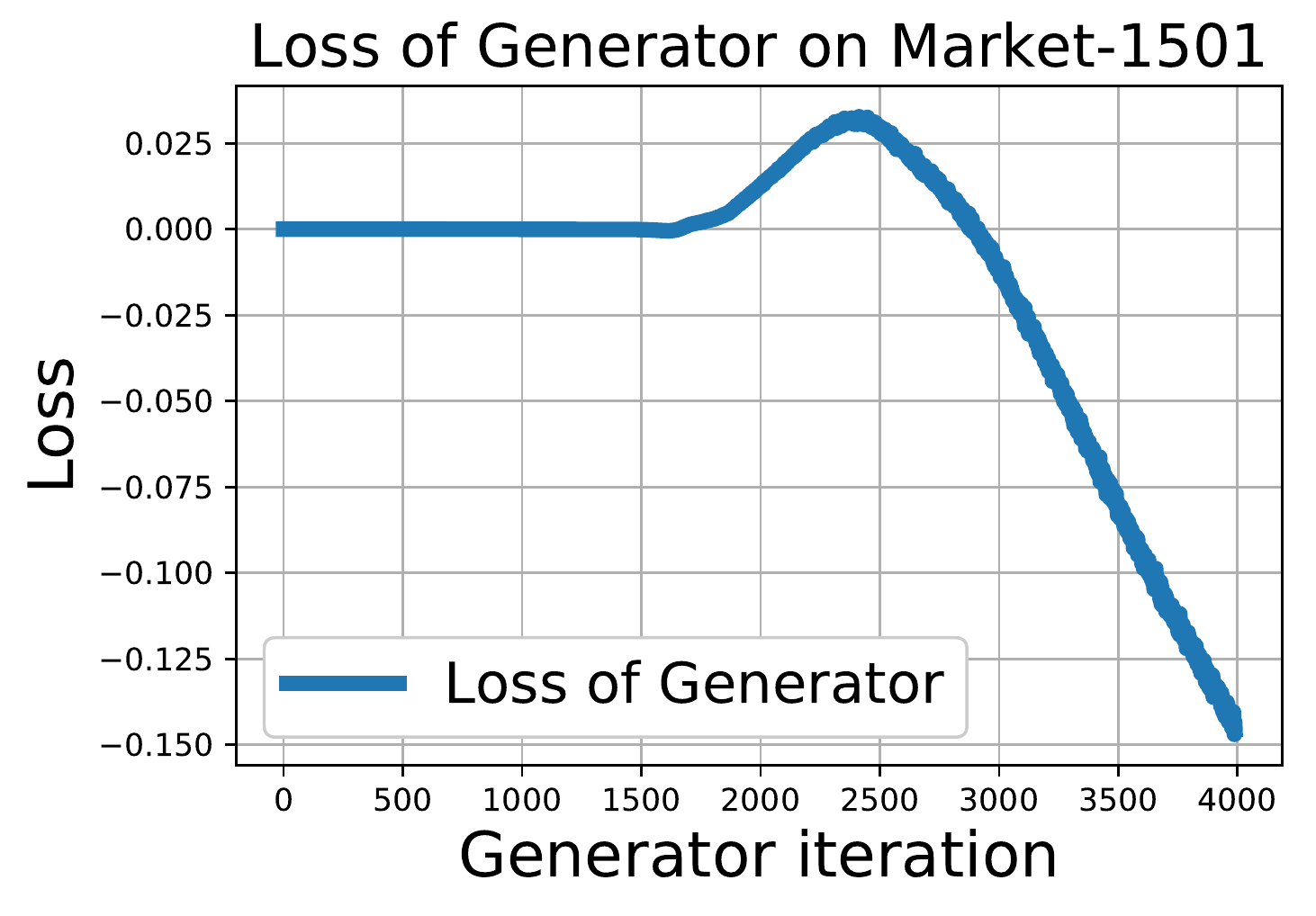}
  %\hspace{-1em}
  \includegraphics[width=0.3\linewidth]{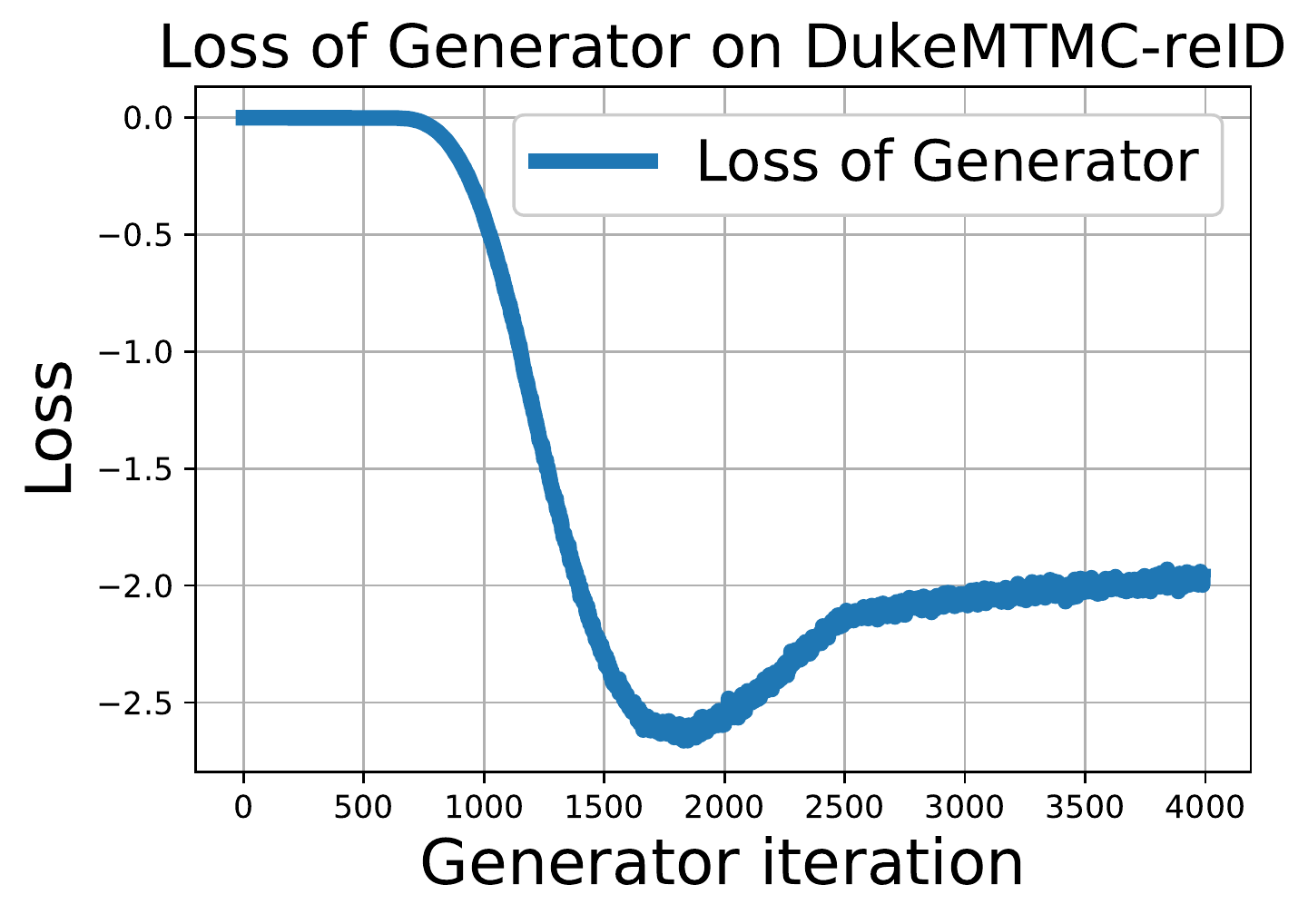}
}

\end{center}
\vspace{-4mm}
\caption{Curves of training losses on CUHK03, Market-1501, and DukeMTMC-reID.}
\vspace{2mm}
\label{fig:losses}
\end{figure}

We first evaluate the efficiency of our method with different bit lengths, since shorter binary codes are more efficient but may cause the descent of accuracies, while longer codes do the opposite. The time of retrieving a query (\textbf{Q. Time}) and the memory storing the gallery features (\textbf{Mem.}) are shown in Fig. \ref{fig:timememory}. As can be seen, the query time and memory consumed by binary codes are far less than those of real-valued features. The matching time and memory of real-valued features increase significantly faster than the binary features with the bit lengths. Besides, we compare the rank-1 matching rates of real-valued and binary features in Table \ref{table:floatbinary}. As we can see from the last two rows of the table, binarized features with more bits (\eg 1024 or 2048) perform only slightly worse than the corresponding real-valued features, which demonstrates that with sufficient capacity, the discriminative information is well preserved by the binary features using our method. It is also noteworthy that even with 2048 bits, our binary features require much less query time and memory than real-valued counterparts.

In addition to the matching time, the time consumed by feature extraction (\textbf{F. Time}) should also be taken into consideration. With the data scale in ReID getting larger, it is necessary to process a large number of queries in a short time. Therefore, we compare the time consumed by extracting features of our method with two state-of-the-art approaches, namely Local Maximal Occurrence Representation (LOMO) \cite{liao2015person} and Hierarchical Gaussian Descriptor (GOG) \cite{Matsukawa_2016_CVPR}, which are widely adopted by traditional ReID methods. As shown in Table~\ref{table:featuretime}, our method extracts features much faster than LOMO and GOG.

% because of the sufficient training samples
Furthermore, we provide the training losses of our framework with 2048 bit length in Fig.~\ref{fig:losses}. We can observe that the losses corresponding to the GAN descend steadily with the proceeding of the training. The triplet loss on Market-1501 is well optimized at every margin value. The triplet losses on CUHK03 and DukeMTMC-reID become fluctuations at certain margin values, nevertheless the losses can reach the steady state by the end of training.

%With the increase of margins, the triplet loss on Market-1501 is well optimized at every margin value, while the triplet losses on CUHK03 and DukeMTMC-reID show instability at certain margin values. Nevertheless, the losses are able to reach the convergence at the end of training.

\begin{table}[t]%\Large
\begin{center}

\caption{Time costs (seconds) of feature extraction for the gallery images.}
\label{table:featuretime}
\vspace{-1mm}
\resizebox{.8\textwidth}{!}{
\begin{tabular}{r|c|c|c}
\hline
\textbf{Method} & \textbf{CUHK03} &\textbf{Market-1501} & \textbf{DukeMTMC-reID}\\
\hline
LOMO\cite{liao2015person}     & 7.50e+00 & 2.96e+02 & 2.65e+02\\
GOG\cite{Matsukawa_2016_CVPR} & 7.10e+02 & 2.80e+04 & 2.51e+04\\
\hline
64-bit ABC+triplet                 & 5.70e$-$01 & 5.93e+00 & 5.05e+00\\
128-bit ABC+triplet                & 5.70e$-$01 & 6.11e+00 & 5.09e+00\\
256-bit ABC+triplet                & 5.71e$-$01 & 6.25e+00 & 5.17e+00\\
512-bit ABC+triplet                & 5.72e$-$01 & 6.30e+00 & 5.19e+00\\
1024-bit ABC+triplet               & 5.73e$-$01 & 6.38e+00 & 5.32e+00\\
2048-bit ABC+triplet               & 5.77e$-$01 & 6.49e+00 & 5.33e+00\\
\hline
\end{tabular}
}
\end{center}
%\vspace{-5mm}
\end{table}

\begin{table}[!b]%\scriptsize
\begin{center}
\newcommand{\tabincell}[2]{\begin{tabular}{@{}#1@{}}#2\end{tabular}}
%\vspace{0.6em}
\caption{Matching rates (\%), mAP (\%), feature extraction time (seconds), average query time (seconds) and memory usage (Million Bytes) for storing gallery data, by comparing with hashing-based ReID methods on CUHK03 (labelled).}
\label{tab:cuhk03hashing}
\vspace{-2mm}
\resizebox{0.95\textwidth}{!}{
\begin{tabular}{c|ccc|c|c|c|c}
\hline
\textbf{Method} & \textbf{Rank 1~} & \textbf{Rank 5~} & \textbf{Rank 20} & \textbf{mAP} & \textbf{F. Time (s)} & \textbf{Q. Time (s)} & \textbf{Mem. (MB)}\\
\hline
\footnotesize
DRSCH \cite{zhang2015bit}             & 22.0 & 48.4 & 81.0 & -    & -        & -        & - \\
DSRH \cite{zhao2015deep}              & 14.4 & 43.4 & 79.2 & -    & -        & -        & - \\
CSBT \cite{ChenJiaxin_2017_CVPR}      & \textcolor{blue}{55.5} & \textcolor{blue}{84.3} & \textcolor{blue}{98.0} & -    & \textcolor{blue}{7.50e+00}& \textcolor{red}{8.07e-04}& \textcolor{red}{1.01e-02}\\
\hline
512-bit ABC+triplet (Ours)         & \textcolor{red}{58.2} & \textcolor{red}{85.7} & \textcolor{red}{98.2} & \textcolor{red}{59.7} & \textcolor{red}{5.72e-01}& \textcolor{red}{8.07e-04}& \textcolor{red}{1.01e-02}\\
\hline
\end{tabular}
}
\end{center}
\vspace{-3mm}
\end{table}

\begin{table}[!b]%\Large%\scriptsize
\begin{center}
\newcommand{\tabincell}[2]{\begin{tabular}{@{}#1@{}}#2\end{tabular}}
%%\vspace{-0.1em}
\caption{Matching rates (\%), mAP (\%), feature extraction time (seconds), average query time (seconds) and memory usage (Million Bytes) for storing gallery data, by comparing with hashing-based ReID methods on Market-1501.}
\label{tab:market1501hashing}
\vspace{-2mm}
\resizebox{0.9\textwidth}{!}{
\begin{tabular}{c|c|c|c|c|c}
\hline
\textbf{Method} & \textbf{Rank 1}& \textbf{mAP} & \textbf{F. Time (s)} & \textbf{Q. Time (s)}& \textbf{Mem. (MB)}\\
\hline
%& DSRH$^{*}$\cite{zhao2015deep}           & -    & -     & -   &  -      & -          \\
CSBT \cite{ChenJiaxin_2017_CVPR}   & \textcolor{blue}{42.9} & \textcolor{blue}{20.3}  & \textcolor{blue}{2.96e+02} & \textcolor{red}{2.54e$-$02} & \textcolor{red}{1.52e+00}\\
\hline
512-bit ABC+triplet (Ours)      & \textcolor{red}{59.3} & \textcolor{red}{43.8} & \textcolor{red}{6.30e+00} & \textcolor{red}{2.54e$-$02} & \textcolor{red}{1.52e+00}\\
\hline
\end{tabular}
}
\end{center}
%\vspace{-5mm}
\end{table}

\subsection{Comparison with Binary Coding based Methods}
\label{subsec:cmphash}

Here we compare our framework with the following state-of-the-art binary coding (hashing) based ReID methods: 1) \textbf{Deep hashing:} Deep Regularized Similarity Comparison Hashing (DRSCH) \cite{zhang2015bit}, Deep Semantic Ranking based Hashing (DSRH) \cite{zhao2015deep}, and 2) \textbf{Non-deep hashing:} Cross-camera Semantic Binary Transformation (CSBT) \cite{ChenJiaxin_2017_CVPR}. Since CSBT has already significantly outperformed other hashing methods (\eg SePH \cite{lin2015semantics}, COSDISH \cite{kang2016column} and SDH \cite{shen2015supervised}) in ReID according to the results reported in \cite{ChenJiaxin_2017_CVPR}, we mainly compare our method with CSBT. The results on different datasets are shown in Tables~\ref{tab:cuhk03hashing} and \ref{tab:market1501hashing}, respectively, where the best performance is highlighted in red and the second best in blue.

%In this section, we compare our framework with the following state-of-the-art \emph{binary coding (hashing)} based methods. The comparing methods include 1) \textbf{Non-deep hashing:} Semantics-preserving Hashing (SePH) \cite{lin2015semantics}, Semantic Correlation Maximization (SCM) \cite{zhang2014large}, Kernel Supervised Hashing (KSH) \cite{liu2012supervised}, FastHash \cite{lin2014fast}, Supervised Discrete Hashing (SDH) \cite{shen2015supervised}, Column Sampling Based Discrete Supervised Hashing (COSDISH) \cite{kang2016column}, Cross-camera Semantic Binary Transformation (CSBT) \cite{ChenJiaxin_2017_CVPR}, and 2) \textbf{Deep hashing:} Deep Regularized Similarity Comparison Hashing (DRSCH) \cite{zhang2015bit}, Deep Semantic Ranking based Hashing (DSRH) \cite{zhao2015deep}. The rank-1 matching rates on CUHK03 with different bit lengths are shown in Table~\ref{tab:cuhk03hashing}, where the best performance is highlighted in red and the second best is highlighted in blue. We also compare our method with the state-of-the-art hashing method CSBT on Market-1501 in both accuracy and efficiency aspects in Table~\ref{tab:market1501hashing}. Note that we adopt the results of the comparing methods reported in \cite{ChenJiaxin_2017_CVPR}.

From Table~\ref{tab:cuhk03hashing}, we can observe that DRSCH and DSRH perform poorly on CUHK03, falling behind the other methods. Our method outperforms the state-of-the-art hashing based ReID method CSBT. The superiority of our method over CSBT becomes greater on the larger dataset, namely Market-1501, as can be seen in Table~\ref{tab:market1501hashing}. We achieve 16.3\% higher in rank-1 accuracy and double the mean average precision (mAP) of CSBT. This is probably because Market-1501 has much more training data than CUHK03, and projection learning based CSBT can hardly handle such amount of data at once. In contrast, our method optimizes the network based on mini-batch learning approaches, which is able to train the model on large amounts of data. Moreover, CSBT requires extracting LOMO features in advance, which is much slower than extracting binary codes using our method.

\subsection{Comparison with the State-of-the-Art Methods}
\label{subsec:cmpstateoftheart}

\begin{table}[b]%\scriptsize
\begin{center}
\newcommand{\tabincell}[2]{\begin{tabular}{@{}#1@{}}#2\end{tabular}}
%%\vspace{0.6em}
\caption{Matching rates (\%), mAP (\%), feature extraction time (seconds), average query time (seconds) and memory usage (Million Bytes) for storing gallery data, by comparing with state-of-the-art non-hashing methods on CUHK03 (detected).}
\label{tab:cuhk03result}
%%\vspace{-1.7em}
\resizebox{0.9\textwidth}{!}{
\begin{tabular}{c|ccc|c|c|c|c}
\hline
\textbf{Method} & \textbf{Rank 1~} & \textbf{Rank 5~} & \textbf{Rank 20} & \textbf{mAP} & \textbf{F. Time (s)} & \textbf{Q. Time (s)} & \textbf{Mem. (MB)}\\
\hline
\footnotesize
DeepReID\cite{li2014deepreid}         & 19.9 & 49.8 & 78.2 & -    & -        & -        & -\\
Improved Deep\cite{ahmed2015improved} & 44.9 & 76.4 & 93.6 & -    & -        & -        & -\\
NSL\cite{Zhang_2016_CVPR}             & 54.7 & 84.8 & 95.2 & -    &  -       &  -       & 2.11e+01\\
Gated CNN\cite{varior2016gated}       & 61.8 & 80.9 &  -   & 51.3 & -        &  -       & - \\
EDM\cite{shi2016embedding}            & 51.9 & 83.6 &  -   & -    & -        & 2.85e$-$02 & \textcolor{blue}{4.90e$-$01}\\
SIR+CIR\cite{wang2016joint}           & 52.2 &  -   &  -   & -    & -        & \textcolor{blue}{1.26e$-$02} & 5.16e$-$01\\
%JSTL\cite{Xiao_2016_CVPR}             &  -   &  -   &  -   & -    & -        & -        & -\\
%DCSL\cite{zhang2016semantics}         &  -   &  -   &  -   & -    & -        & -        & 5.18e-01\\
%Quadruplet\cite{Chen_2017_CVPR}       & -    & -    &  -   &  -   &  -       & -        & -\\
Re-ranking\cite{Zhong_2017_CVPR}      & 58.5 & -    & -    & \textcolor{red}{64.7}& \textcolor{blue}{7.50e+00}& -        & 1.03e+02\\
SSM\cite{Bai_2017_CVPR}               & 72.7 & 92.4 &  -   & -    & 7.18e+02 & 6.00e$-$01 & 2.08e+02\\
Part-aligned\cite{Zhao_2017_ICCV}     & \textcolor{red}{81.6}& \textcolor{red}{97.3}& \textcolor{red}{99.5}&  -   &  -       & -        & -\\
MuDeep\cite{Qian_2017_ICCV}           & 75.6 & 94.4 & -    &  -   &  -       & -        & -\\
PDC\cite{Su_2017_ICCV}                & \textcolor{blue}{78.3}& \textcolor{blue}{94.8}& \textcolor{blue}{98.4}&  -   &  -       & -        & -\\
\hline
ABC+triplet (Ours)         & 68.1 & 90.3 & 98.3 & \textcolor{blue}{61.6}& \textcolor{red}{5.77e$-$01}& \textcolor{red}{1.85e$-$03}& \textcolor{red}{1.22e$-$01}\\
\hline
\end{tabular}
}

%\scriptsize
%\textbf{(`*': Experimental results with the optimal bit length are adopted. `-': The source codes or implemental details are not available.)}
\end{center}
%\vspace{-5mm}
\end{table}

\begin{table}[t]%\Large%\scriptsize
\begin{center}
\newcommand{\tabincell}[2]{\begin{tabular}{@{}#1@{}}#2\end{tabular}}
%%\vspace{-0.1em}
\caption{Matching rates (\%), mAP (\%), feature extraction time (seconds), average query time (seconds) and memory usage (Million Bytes) for storing gallery data, by comparing with state-of-the-art non-hashing methods on Market-1501.}
\label{tab:market1501result}
\vspace{-2mm}
\resizebox{0.8\textwidth}{!}{
\begin{tabular}{c|c|c|c|c|c}
\hline
\textbf{Method} & \textbf{Rank 1}& \textbf{mAP} & \textbf{F. Time (s)} & \textbf{Q. Time (s)}& \textbf{Mem. (MB)}\\
\hline
SDALF\cite{farenzena2010person}    & 20.5 & 8.2  &  -       & 2.53e+02 & 9.31e+01  \\
KISSME\cite{koestinger2012large}   & 40.5 & 19.0 &  -       & -        &  -         \\
eSDC\cite{zhao2013unsupervised}    & 33.5 & 13.5 & 1.58e+04 & 7.47e+02 & 1.45e+03  \\
BoW(best)\cite{zheng2015scalable}  & 44.4 & 21.9 &  -       & \textcolor{blue}{2.08e+00} & \textcolor{blue}{1.54e+01}  \\
LOMO+NSL\cite{Zhang_2016_CVPR}     & 55.4 & 29.9 & \textcolor{blue}{2.96e+02} & -        & 4.06e+03\\
Gated CNN\cite{varior2016gated}    & 65.9 & 39.6 &  -       & -        & 4.32e+01  \\
%SSDAL\cite{su2016deep}             & 39.4 & 19.6 &  -       & -        &  -         \\
SCSP\cite{Chen_2016_CVPR}          & 51.9 & 26.4 &  -       & -        & 1.21e+02  \\
%P2S\cite{ZhouSanping_2017_CVPR}    & 70.7 & 44.3 &  -       & -        &  -         \\
SpindleNet\cite{Zhao_2017_CVPR}    & 76.9 & -    &  -       & -        &  -         \\
%CADL\cite{Lin_2017_CVPR}           & 73.8 & 47.1 &  -       & -        &  -         \\
Re-ranking\cite{Zhong_2017_CVPR}   & 77.1 & \textcolor{blue}{63.6} & \textcolor{blue}{2.96e+02} & -        & 4.06e+03\\
SSM\cite{Bai_2017_CVPR}            & \textcolor{blue}{82.2} & \textcolor{red}{68.8} & 2.83e+04 & 1.68e+02 & 8.21e+03\\
Part-aligned\cite{Zhao_2017_ICCV}  & 81.0 & 63.4 &  -       & -        &  -         \\
PDC\cite{Su_2017_ICCV}             & \textcolor{red}{84.1} & 63.4 &  -       & -        &  -         \\
\hline
ABC+triplet (Ours)      & 73.5 & 52.9 & \textcolor{red}{6.49e+00} & \textcolor{red}{7.32e$-$02} & \textcolor{red}{4.82e+00}\\
\hline
\end{tabular}
}
\end{center}
\vspace{-3mm}
\end{table}

\begin{table}[t]%\Large%\scriptsize
\begin{center}
\newcommand{\tabincell}[2]{\begin{tabular}{@{}#1@{}}#2\end{tabular}}
%%\vspace{-0.1em}
\caption{Matching rates (\%), mAP (\%), feature extraction time (seconds), average query time (seconds) and memory usage (Million Bytes) for storing gallery data, by comparing with state-of-the-art non-hashing methods on DukeMTMC-reID.}
\label{tab:dukeresult}
\vspace{-2mm}
\resizebox{0.8\textwidth}{!}{
\begin{tabular}{c|c|c|c|c|c}
\hline
\textbf{Method} & \textbf{Rank 1} & \textbf{mAP} & \textbf{F. Time (s)} & \textbf{Q. Time (s)} & \textbf{Mem. (MB)}\\
\hline
BoW+KISSME\cite{zheng2015scalable}    & 25.l & 12.1 & -        &  - &  -  \\
LOMO+XQDA\cite{liao2015person}        & 30.8 & 17.0 & \textcolor{blue}{2.65e+02} &  - & \textcolor{blue}{3.62e+03}\\
Basel.+LSRO\cite{zheng2017unlabeled}  & 67.7 & 47.1 & -        &  - &  -  \\
Basel.+OIM\cite{xiao2017joint}        & 68.1 &  -   & -        &  - &  -  \\
SVDNet\cite{sun_iccv2017unlabeled}    & \textcolor{blue}{76.7} & \textcolor{red}{56.8} & -        &  - &  -  \\
\hline
ABC+triplet (Ours)      & \textcolor{red}{77.6} & \textcolor{blue}{47.9} & \textcolor{red}{5.33e+00} & \textcolor{red}{6.58e$-$02} & \textcolor{red}{4.31e+00}\\
\hline
\end{tabular}
}
\end{center}
%\vspace{-5mm}
\end{table}

% Quadruplet Net \cite{Chen_2017_CVPR},DCSL \cite{zhang2016semantics}, One-shot learning \cite{Bak_2017_CVPR}

%SSDAL \cite{su2016deep}, P2S \cite{ZhouSanping_2017_CVPR}, CADL \cite{Lin_2017_CVPR},
We also compare our method (2048-bit ABC+triplet) with the state-of-the-art \emph{non-hashing} ReID methods. The methods mainly include: 1)~\textbf{Deep Learning based methods}, such as DeepReID \cite{li2014deepreid}, Improved Deep \cite{ahmed2015improved}, SIR+CIR \cite{wang2016joint}, EDM \cite{shi2016embedding}, Gated CNN \cite{varior2016gated}, Spindle Net \cite{Zhao_2017_CVPR}, SVDNet \cite{sun_iccv2017unlabeled}, Deeply-learned Part-Aligned Representation \cite{Zhao_2017_ICCV}, Multi-scale Deep Learning Architectures \cite{Qian_2017_ICCV}, and Pose-Driven Deep Convolutional Model \cite{Su_2017_ICCV}; 2)~\textbf{Metric learning based methods}, such as KISSME \cite{koestinger2012large}, XQDA \cite{liao2015person}, and NSL \cite{Zhang_2016_CVPR}; 3)~\textbf{Local Patch Matching based methods}, such as SDC \cite{zhao2013unsupervised} and BoW \cite{zheng2015scalable}; 4)~Other ReID methods.

The comparison results on the three datasets are shown in Tables~\ref{tab:cuhk03result}, \ref{tab:market1501result}, and \ref{tab:dukeresult}, respectively. Our framework achieves competitive matching accuracies compared to the state-of-the-art methods, which adopt high-dimensional real-value features. It is also obviously that our framework not only outperforms many existing non-hashing approaches, but also achieves significant advantages in terms of the matching efficiencies. The advantages are more outstanding if the gallery set contains more samples. For instance, the query time of ABC is at least dozens of times faster than the non-hashing methods on Market-1501 with 19,732 gallery samples. Several methods adopt LOMO, which represents images as 26960-dim real-valued features. Differently, our method just represents images as 2048-bit binary codes which requires far less memory.

\subsection{Effects of Different Network Settings}
\label{subsec:ablation}

In this section, we embed the proposed ABC into different similarity measuring networks and evaluate the performance under different settings. We first evaluate two types of networks widely used to measure similarity, namely siamese network \cite{shi2016embedding,yi2014deep} and triplet network. A siamese network receives a pair of images and minimizes the distance between images if they are from the same class and maximizes the distance if they have different labels. The siamese network evaluated in our experiments adopts the same ResNet-50 backbone model as the triplet network and employs the contractive loss to measure similarity.

From Table~\ref{tab:ablation}, we can observe that adopting siamese network performs worse than triplet network. This is because the loss of the siamese network is too strict, \ie enforcing images of one identity to be projected onto a single point in the subspace. Differently, the triplet loss allows the images from one person to lie on a manifold, while enforcing larger distances between different persons' images. We can also observe that embedding the ABC into a triplet network achieves better results than into a siamese network.

\begin{table}[t]%\scriptsize
\begin{center}
\newcommand{\tabincell}[2]{\begin{tabular}{@{}#1@{}}#2\end{tabular}}
%%\vspace{0.6em}
\caption{Matching rates (\%) and mAP (\%) of the proposed ABC embedded networks in terms of different settings.}
\label{tab:ablation}
%%\vspace{-1.7em}
\resizebox{0.99\textwidth}{!}{
\begin{tabular}{c|cc|cc|cc|cc}
\hline
\multirow{2}{*}{\tabincell{c}{\textbf{Settings}}}
& \multicolumn{2}{c|}{ \textbf{CUHK03 (labelled)} }& \multicolumn{2}{c|}{ \textbf{CUHK03 (detected)} } & \multicolumn{2}{c|}{ \textbf{Market-1501} } & \multicolumn{2}{c}{ \textbf{DukeMTMC-reID} } \\
\cline{2-9}
& ~~Rank 1~~ & mAP & ~~~Rank 1~~~ & mAP & Rank 1 & mAP & ~~~Rank 1~~~ & mAP\\
\hline
2048-dim siamese              & 62.0 & 59.4 & 61.5 & 56.1 & 67.2 & 48.8 & 75.0 & 43.8\\
2048-dim triplet              & 70.8 & 66.6 & 69.7 & 63.1 & 75.2 & 54.1 & 82.0 & 48.8\\
2048-bit ABC+siamese          & 49.2 & 35.4 & 45.6 & 31.2 & 52.7 & 21.8 & 56.9 & 29.7\\
2048-bit ABC+triplet          & 52.3 & 43.7 & 50.9 & 38.1 & 55.8 & 27.5 & 60.3 & 27.6\\
2048-bit ABC+siamese+$\ell_2$       & 61.7 & 60.4 & 61.6 & 60.2 & 65.7 & 48.1 & 70.9 & 41.7\\
2048-bit ABC+triplet+$\ell_2$       & 68.8 & 64.5 & 68.1 & 61.6 & 73.5 & 52.9 & 77.6 & 47.9\\
\hline
\end{tabular}
}
\end{center}
\vspace{-4mm}
\end{table}

\begin{figure}[t]
\begin{center}
\subfigure[CUHK03]{ \label{subfig:cuhk03finetune}
%\fbox{\rule{0pt}{1in} \rule{0.29\linewidth}{0pt}}
  \includegraphics[width=0.325\linewidth]{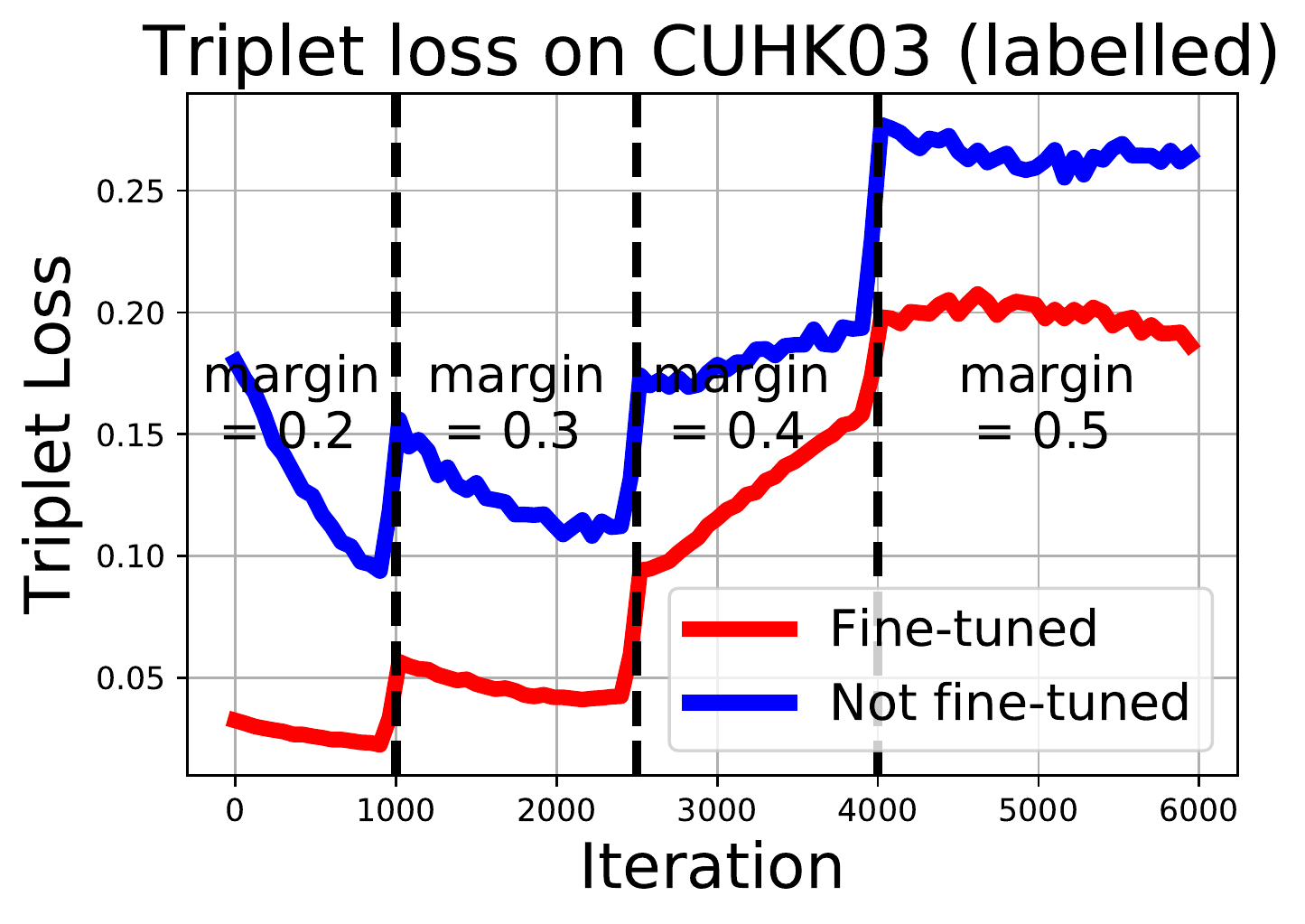}
}
\hspace{-4mm}
\subfigure[Market-1501]{ \label{subfig:marketfinetune}
% \fbox{\rule{0pt}{1in} \rule{0.29\linewidth}{0pt}}
  \includegraphics[width=0.325\linewidth]{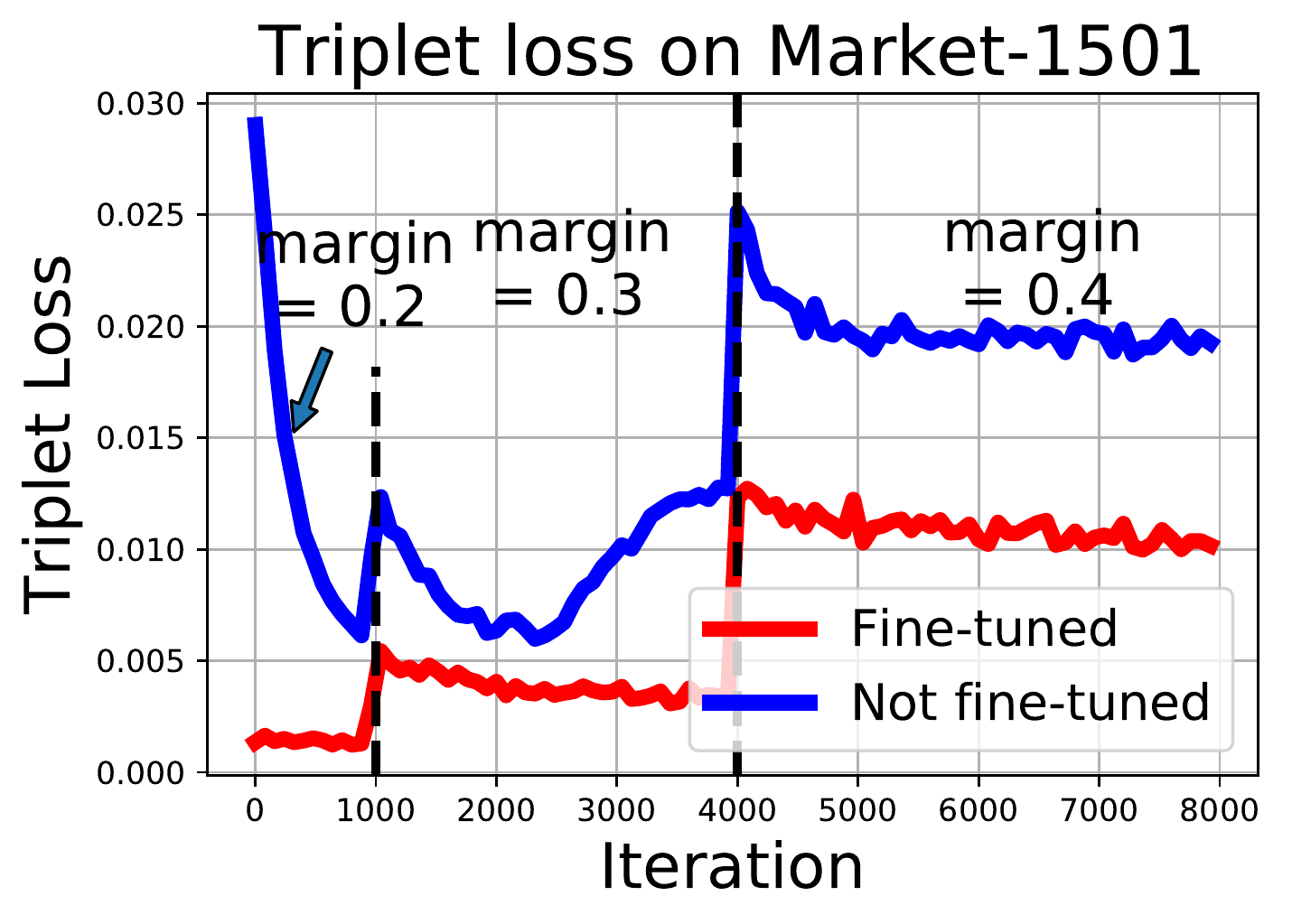}
}
\hspace{-4mm}
\subfigure[DukeMTMC-reID]{ \label{subfig:dukefinetune}
%\fbox{\rule{0pt}{1in} \rule{0.29\linewidth}{0pt}}
  \includegraphics[width=0.325\linewidth]{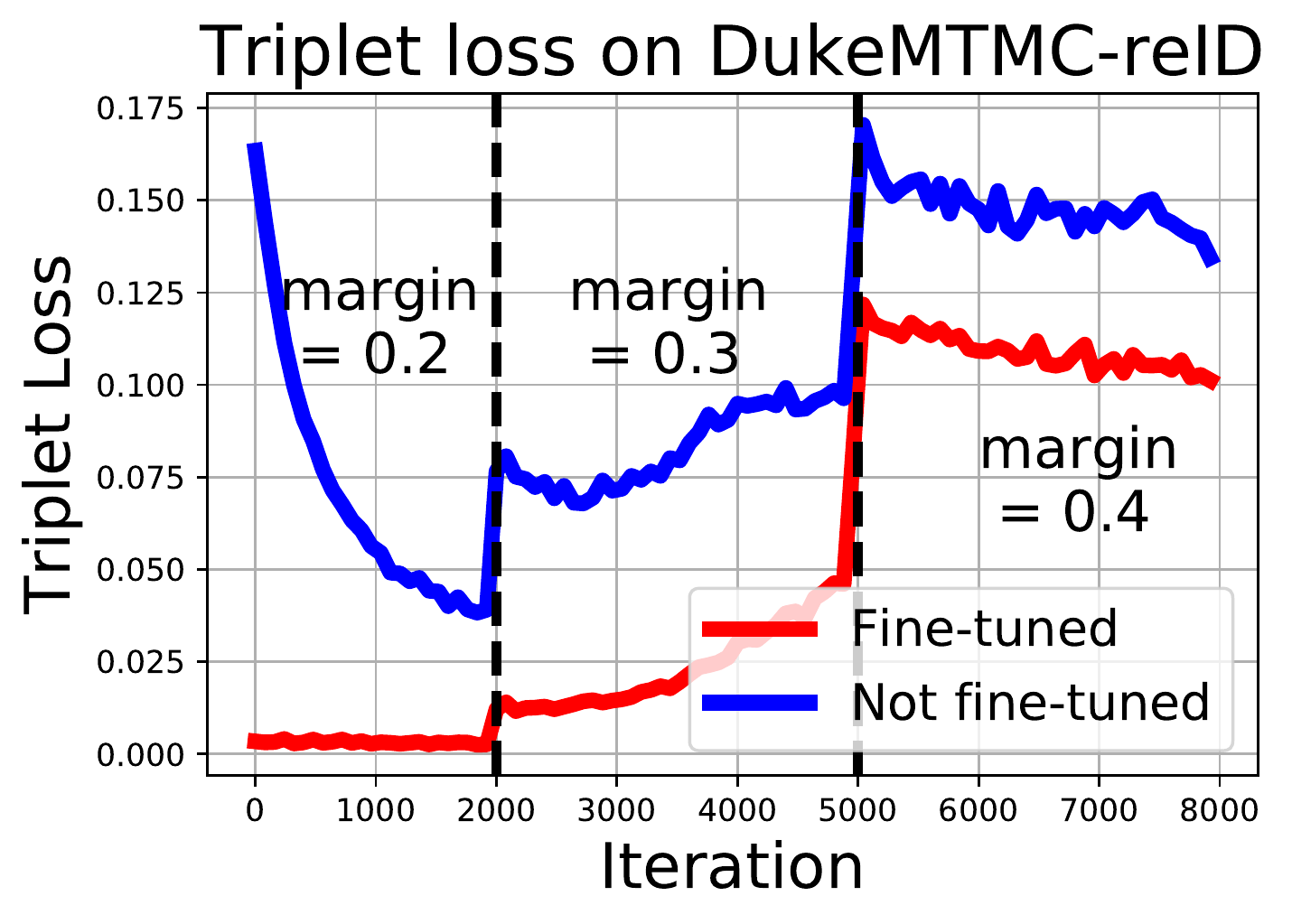}
}
\end{center}
\vspace{-3mm}
   \caption{Triplet losses with and without the fine-tuning phase on pre-trained models on CUHK03, Market-1501, and DukeMTMC-reID. The bit length is 2,048, and the backbone model is ResNet-50.}
\label{fig:iffinetune}
\vspace{3.6mm}
\end{figure}

As explained in Section \ref{subsec:dalbc}, we normalize both the generated features and binary codes to the same scale to eliminate the conflict between two modules. Here we also compare the networks with and without the $\ell_2$ normalization. As can be seen from Table~\ref{tab:ablation}, the performance is significantly improved after the normalization.

In addition, we evaluate the effect of fine-tuning on the three datasets. Fig.~\ref{fig:iffinetune} shows that the triplet losses with fine-tuning converge faster than those without fine-tuning. Since the employed ResNet-50 network is pre-trained on ImageNet, it already captures a variety of useful image features. Fine-tuning the network further enables learning features specialized for person representation more efficiently than training the network from scratch.

%-------------------------------------------------------------------------
\section{Conclusion}
\label{sec:conclusion}

In this work, the adversarial binary coding (ABC) framework for efficient person re-identification was proposed, which could generate discriminative and efficient binary features from pedestrian images. Specifically, our ABC trained a discriminator network to distinguish the real-valued features from binary ones, in order to guide the feature extractor network to generate features in binary form under the Wasserstein loss. The ABC framework was further embedded into a deep triplet network to preserve the semantic information of binary features for the ReID task. Extensive experiments on three large-scale ReID datasets showed that our method outperformed the state-of-the-art hashing based ReID approaches, and was competitive to the state-of-the-art non-hashing ReID approaches, whilst reducing time and memory costs significantly. Considering that the triplet network has been overtaken by other network architectures proposed more recently, one possible improvement of this work in future is to explore the combinations of the ABC framework and other more sophisticated similarity measuring frameworks.

\bibliographystyle{splncs}
\bibliography{reid_short}

\begin{thebibliography}{10}

\bibitem{farenzena2010person}
Farenzena, M., Bazzani, L., Perina, A., Murino, V., Cristani, M.:
\newblock Person re-identification by symmetry-driven accumulation of local
  features.
\newblock In: CVPR. (2010)

\bibitem{liao2015person}
Liao, S., Hu, Y., Zhu, X., Li, S.Z.:
\newblock Person re-identification by local maximal occurrence representation
  and metric learning.
\newblock In: CVPR. (2015)

\bibitem{Matsukawa_2016_CVPR}
Matsukawa, T., Okabe, T., Suzuki, E., Sato, Y.:
\newblock Hierarchical gaussian descriptor for person re-identification.
\newblock In: CVPR. (2016)

\bibitem{liu2015spatio}
Liu, K., Ma, B., Zhang, W., Huang, R.:
\newblock A spatio-temporal appearance representation for viceo-based
  pedestrian re-identification.
\newblock In: ICCV. (2015)

\bibitem{shi2015transferring}
Shi, Z., Hospedales, T.M., Xiang, T.:
\newblock Transferring a semantic representation for person re-identification
  and search.
\newblock In: CVPR. (2015)

\bibitem{wang2016joint}
Wang, F., Zuo, W., Lin, L., Zhang, D., Zhang, L.:
\newblock Joint learning of single-image and cross-image representations for
  person re-identification.
\newblock In: CVPR. (2016)

\bibitem{zheng2016person}
Zheng, L., Yang, Y., Hauptmann, A.G.:
\newblock Person re-identification: Past, present and future.
\newblock arXiv preprint arXiv:1610.02984 (2016)

\bibitem{zheng2016learning}
Zheng, F., Shao, L.:
\newblock Learning cross-view binary identities for fast person
  re-identification.
\newblock In: IJCAI. (2016)  2399--2406

\bibitem{ChenJiaxin_2017_CVPR}
Chen, J., Wang, Y., Qin, J., Liu, L., Shao, L.:
\newblock Fast person re-identification via cross-camera semantic binary
  transformation.
\newblock In: CVPR. (2017)

\bibitem{weiss2009spectral}
Weiss, Y., Torralba, A., Fergus, R.:
\newblock Spectral hashing.
\newblock In: NIPS. (2009)

\bibitem{goodfellow2014generative}
Goodfellow, I., Pouget-Abadie, J., Mirza, M., Xu, B., Warde-Farley, D., Ozair,
  S., Courville, A., Bengio, Y.:
\newblock Generative adversarial nets.
\newblock In: Advances in neural information processing systems. (2014)
  2672--2680

\bibitem{radford2015unsupervised}
Radford, A., Metz, L., Chintala, S.:
\newblock Unsupervised representation learning with deep convolutional
  generative adversarial networks.
\newblock arXiv preprint arXiv:1511.06434 (2015)

\bibitem{li2014deepreid}
Li, W., Zhao, R., Xiao, T., Wang, X.:
\newblock Deepreid: Deep filter pairing neural network for person
  re-identification.
\newblock In: CVPR. (2014)

\bibitem{zheng2015scalable}
Zheng, L., Shen, L., Tian, L., Wang, S., Wang, J., Tian, Q.:
\newblock Scalable person re-identification: A benchmark.
\newblock In: ICCV. (2015)

\bibitem{zheng2017unlabeled}
Zheng, Z., Zheng, L., Yang, Y.:
\newblock Unlabeled samples generated by gan improve the person
  re-identification baseline in vitro.
\newblock In: ICCV. (2017)

\bibitem{koestinger2012large}
Koestinger, M., Hirzer, M., Wohlhart, P., Roth, P.M., Bischof, H.:
\newblock Large scale metric learning from equivalence constraints.
\newblock In: CVPR. (2012)

\bibitem{pedagadi2013local}
Pedagadi, S., Orwell, J., Velastin, S., Boghossian, B.:
\newblock Local fisher discriminant analysis for pedestrian re-identification.
\newblock In: CVPR. (2013)

\bibitem{prosser2010person}
Prosser, B., Zheng, W.S., Gong, S., Xiang, T., Mary, Q.:
\newblock Person re-identification by support vector ranking.
\newblock In: BMVC. Volume~2. (2010)

\bibitem{lisanti2015person}
Lisanti, G., Masi, I., Bagdanov, A.D., Del~Bimbo, A.:
\newblock Person re-identification by iterative re-weighted sparse ranking.
\newblock IEEE TPAMI \textbf{37}(8) (2015)  1629--1642

\bibitem{lan2016quaternionic}
Lan, R., Zhou, Y., Tang, Y.Y.:
\newblock Quaternionic local ranking binary pattern: A local descriptor of
  color images.
\newblock IEEE TIP \textbf{25}(2) (2016)  566--579

\bibitem{ahmed2015improved}
Ahmed, E., Jones, M., Marks, T.K.:
\newblock An improved deep learning architecture for person re-identification.
\newblock In: CVPR. (2015)

\bibitem{Xiao_2016_CVPR}
Xiao, T., Li, H., Ouyang, W., Wang, X.:
\newblock Learning deep feature representations with domain guided dropout for
  person re-identification.
\newblock In: CVPR. (2016)

\bibitem{ZhouSanping_2017_CVPR}
Zhou, S., Wang, J., Wang, J., Gong, Y., Zheng, N.:
\newblock Point to set similarity based deep feature learning for person
  re-identification.
\newblock In: CVPR. (2017)

\bibitem{Lin_2017_CVPR}
Lin, J., Ren, L., Lu, J., Feng, J., Zhou, J.:
\newblock Consistent-aware deep learning for person re-identification in a
  camera network.
\newblock In: CVPR. (2017)

\bibitem{Panda_2017_CVPR}
Panda, R., Bhuiyan, A., Murino, V., Roy-Chowdhury, A.K.:
\newblock Unsupervised adaptive re-identification in open world dynamic camera
  networks.
\newblock In: CVPR. (2017)

\bibitem{Chen_2017_CVPR}
Chen, W., Chen, X., Zhang, J., Huang, K.:
\newblock Beyond triplet loss: A deep quadruplet network for person
  re-identification.
\newblock In: CVPR. (2017)

\bibitem{yi2014deep}
Yi, D., Lei, Z., Liao, S., Li, S.Z.:
\newblock Deep metric learning for person re-identification.
\newblock In: ICPR. (2014)

\bibitem{shi2016embedding}
Shi, H., Yang, Y., Zhu, X., Liao, S., Lei, Z., Zheng, W., Li, S.Z.:
\newblock Embedding deep metric for person re-identification: A study against
  large variations.
\newblock In: ECCV. (2016)

\bibitem{varior2016gated}
Varior, R.R., Haloi, M., Wang, G.:
\newblock Gated siamese convolutional neural network architecture for human
  re-identification.
\newblock In: ECCV. (2016)

\bibitem{chen2016deep}
Chen, S.Z., Guo, C.C., Lai, J.H.:
\newblock Deep ranking for person re-identification via joint representation
  learning.
\newblock IEEE TIP \textbf{25}(5) (2016)  2353--2367

\bibitem{cheng2016person}
Cheng, D., Gong, Y., Zhou, S., Wang, J., Zheng, N.:
\newblock Person re-identification by multi-channel parts-based cnn with
  improved triplet loss function.
\newblock In: CVPR. (2016)

\bibitem{zhao2015deep}
Zhao, F., Huang, Y., Wang, L., Tan, T.:
\newblock Deep semantic ranking based hashing for multi-label image retrieval.
\newblock In: CVPR. (2015)

\bibitem{zhang2015bit}
Zhang, R., Lin, L., Zhang, R., Zuo, W., Zhang, L.:
\newblock Bit-scalable deep hashing with regularized similarity learning for
  image retrieval and person re-identification.
\newblock IEEE TIP \textbf{24}(12) (2015)  4766--4779

\bibitem{huang2017stacked}
Huang, X., Li, Y., Poursaeed, O., Hopcroft, J., Belongie, S.:
\newblock Stacked generative adversarial networks.
\newblock In: CVPR. (2017)

\bibitem{huang2017beyond}
Huang, R., Zhang, S., Li, T., He, R.,  et~al.:
\newblock Beyond face rotation: Global and local perception gan for
  photorealistic and identity preserving frontal view synthesis.
\newblock arXiv preprint arXiv:1704.04086 (2017)

\bibitem{mao2017least}
Mao, X., Li, Q., Xie, H., Lau, R.Y., Wang, Z., Smolley, S.P.:
\newblock Least squares generative adversarial networks.
\newblock In: ICCV. (2017)

\bibitem{zhu2017unpaired}
Zhu, J.Y., Park, T., Isola, P., Efros, A.A.:
\newblock Unpaired image-to-image translation using cycle-consistent
  adversarial networks.
\newblock In: ICCV. (2017)

\bibitem{pix2pix_2017_CVPR}
Isola, P., Zhu, J.Y., Zhou, T., Efros, A.A.:
\newblock Image-to-image translation with conditional adversarial networks.
\newblock In: CVPR. (2017)

\bibitem{makhzani2015adversarial}
Makhzani, A., Shlens, J., Jaitly, N., Goodfellow, I., Frey, B.:
\newblock Adversarial autoencoders.
\newblock arXiv preprint arXiv:1511.05644 (2015)

\bibitem{donahue2016adversarial}
Donahue, J., Kr{\"a}henb{\"u}hl, P., Darrell, T.:
\newblock Adversarial feature learning.
\newblock arXiv preprint arXiv:1605.09782 (2016)

\bibitem{dumoulin2016adversarially}
Dumoulin, V., Belghazi, I., Poole, B., Mastropietro, O., Lamb, A., Arjovsky,
  M., Courville, A.:
\newblock Adversarially learned inference.
\newblock arXiv preprint arXiv:1606.00704 (2016)

\bibitem{arjovsky2017wasserstein}
Arjovsky, M., Chintala, S., Bottou, L.:
\newblock Wasserstein gan.
\newblock arXiv preprint arXiv:1701.07875 (2017)

\bibitem{gulrajani2017improved}
Gulrajani, I., Ahmed, F., Arjovsky, M., Dumoulin, V., Courville, A.:
\newblock Improved training of wasserstein gans.
\newblock arXiv preprint arXiv:1704.00028 (2017)

\bibitem{qiu2017deep}
Qiu, Z., Pan, Y., Yao, T., Mei, T.:
\newblock Deep semantic hashing with generative adversarial networks.
\newblock In: SIGIR, ACM (2017)

\bibitem{song2018ganbinary}
Song, J.:
\newblock Binary generative adversarial networks for image retrieval.
\newblock In: AAAI. (2018)

\bibitem{arjovsky2017towards}
Arjovsky, M., Bottou, L.:
\newblock Towards principled methods for training generative adversarial
  networks.
\newblock arXiv preprint arXiv:1701.04862 (2017)

\bibitem{he2016deep}
He, K., Zhang, X., Ren, S., Sun, J.:
\newblock Deep residual learning for image recognition.
\newblock In: CVPR. (2016)

\bibitem{balntas2016learning}
Balntas, V., Riba, E., Ponsa, D., Mikolajczyk, K.:
\newblock Learning local feature descriptors with triplets and shallow
  convolutional neural networks.
\newblock In: BMVC. (2016)

\bibitem{imagenet2009}
Deng, J., Dong, W., Socher, R., Li, L.J., Li, K., Fei-Fei, L.:
\newblock {ImageNet: A Large-Scale Hierarchical Image Database}.
\newblock In: CVPR. (2009)

\bibitem{ristani2016MTMC}
Ristani, E., Solera, F., Zou, R., Cucchiara, R., Tomasi, C.:
\newblock Performance measures and a data set for multi-target, multi-camera
  tracking.
\newblock In: ECCV Workshop. (2016)

\bibitem{lin2015semantics}
Lin, Z., Ding, G., Hu, M., Wang, J.:
\newblock Semantics-preserving hashing for cross-view retrieval.
\newblock In: CVPR. (2015)

\bibitem{kang2016column}
Kang, W.C., Li, W.J., Zhou, Z.H.:
\newblock Column sampling based discrete supervised hashing.
\newblock In: AAAI. (2016)  1230--1236

\bibitem{shen2015supervised}
Shen, F., Shen, C., Liu, W., Shen, H.T.:
\newblock Supervised discrete hashing.
\newblock In: CVPR. Volume~2. (2015) ~5

\bibitem{Zhang_2016_CVPR}
Zhang, L., Xiang, T., Gong, S.:
\newblock Learning a discriminative null space for person re-identification.
\newblock In: CVPR. (2016)

\bibitem{Zhong_2017_CVPR}
Zhong, Z., Zheng, L., Cao, D., Li, S.:
\newblock Re-ranking person re-identification with k-reciprocal encoding.
\newblock In: CVPR. (2017)

\bibitem{Bai_2017_CVPR}
Bai, S., Bai, X., Tian, Q.:
\newblock Scalable person re-identification on supervised smoothed manifold.
\newblock In: CVPR. (2017)

\bibitem{Zhao_2017_ICCV}
Zhao, L., Li, X., Zhuang, Y., Wang, J.:
\newblock Deeply-learned part-aligned representations for person
  re-identification.
\newblock In: ICCV. (2017)

\bibitem{Qian_2017_ICCV}
Qian, X., Fu, Y., Jiang, Y.G., Xiang, T., Xue, X.:
\newblock Multi-scale deep learning architectures for person re-identification.
\newblock In: ICCV. (2017)

\bibitem{Su_2017_ICCV}
Su, C., Li, J., Zhang, S., Xing, J., Gao, W., Tian, Q.:
\newblock Pose-driven deep convolutional model for person re-identification.
\newblock In: ICCV. (2017)

\bibitem{zhao2013unsupervised}
Zhao, R., Ouyang, W., Wang, X.:
\newblock Unsupervised salience learning for person re-identification.
\newblock In: CVPR. (2013)

\bibitem{Chen_2016_CVPR}
Chen, D., Yuan, Z., Chen, B., Zheng, N.:
\newblock Similarity learning with spatial constraints for person
  re-identification.
\newblock In: CVPR. (2016)

\bibitem{Zhao_2017_CVPR}
Zhao, H., Tian, M., Sun, S., Shao, J., Yan, J., Yi, S., Wang, X., Tang, X.:
\newblock Spindle net: Person re-identification with human body region guided
  feature decomposition and fusion.
\newblock In: CVPR. (2017)

\bibitem{xiao2017joint}
Xiao, T., Li, S., Wang, B., Lin, L., Wang, X.:
\newblock Joint detection and identification feature learning for person
  search.
\newblock In: CVPR. (2017)

\bibitem{sun_iccv2017unlabeled}
Sun, Y., Zheng, L., Deng, W., Wang, S.:
\newblock Svdnet for pedestrian retrieval.
\newblock In: ICCV. (2017)

\end{thebibliography}
\end{document}